\documentclass[12pt]{article}

\usepackage{graphicx}
\usepackage{multicol}
\usepackage{amsmath}
\usepackage{amssymb}

\usepackage{geometry}
\usepackage{color}
\usepackage{bbold}
\usepackage{lscape}
\usepackage{setspace}
\usepackage{array}
\usepackage{multirow}
\usepackage{booktabs}
\usepackage{geometry}
\usepackage[font=small,labelfont=bf]{caption}
\usepackage{subfig}
\usepackage{fancybox}
\usepackage{enumitem}
\usepackage{tabularx}
\usepackage{float}
\usepackage{ifthen}
\usepackage{amsfonts}
\usepackage{amsthm}
\usepackage{eurosym}
\usepackage{setspace}
\usepackage{array}
\usepackage{multirow}
\usepackage{hyperref}
\usepackage{footmisc}

\usepackage{dsfont}
\usepackage[table]{xcolor}
\usepackage{lscape}
\usepackage{savefnmark}

\newcommand{\blanco}[1]{}

\makeatletter
\def\maxwidth{ %
  \ifdim\Gin@nat@width>\linewidth
    \linewidth
  \else
    \Gin@nat@width
  \fi
}
\makeatother

\definecolor{fgcolor}{rgb}{0.345, 0.345, 0.345}

\usepackage{framed}
\makeatletter
 {\par\unskip\endMakeFramed%
 \at@end@of@kframe}
\makeatother

\definecolor{shadecolor}{rgb}{.97, .97, .97}
\definecolor{messagecolor}{rgb}{0, 0, 0}
\definecolor{warningcolor}{rgb}{1, 0, 1}
\definecolor{errorcolor}{rgb}{1, 0, 0}
\usepackage{alltt}

\usepackage{mathptmx}

\usepackage{graphics}

\usepackage{natbib} 

\usepackage{tikz}
\usetikzlibrary{trees}
\usepackage{colortbl}
\usepackage{color}
\usepackage{booktabs}
\usepackage{hyperref}
\usepackage{amsmath}
\usepackage{amssymb}
\usepackage{float}
\usepackage{subfig}

\definecolor{lightgray}{rgb}{0.75, 0.75, 0.75}

\def\lc{\left\lfloor}   
\def\rc{\right\rfloor}

\begin{document}

\title{Hybrid Machine Learning Forecasts for the UEFA EURO 2020}
%\title{A combined ranking-based machine learning approach to forecast the FIFA Women's World Cup~2019}

\author{A. Groll
\thanks{Department of Statistics, TU Dortmund University, Vogelpothsweg 87, 44227 Dortmund, Germany, \emph{groll@statistik.tu-dortmund.de}}
\and L. M. Hvattum
\thanks{Molde University College, Molde, Norway, \emph{Lars.M.Hvattum@himolde.no}}
\and C. Ley
\thanks{Faculty of Sciences, Department of Applied Mathematics, Computer Science and Statistics, Ghent University, Krijgslaan 281, 9000 Gent,  Belgium, \emph{Christophe.Ley@UGent.be}}
\and F. Popp
\thanks{Department of Statistics, TU Dortmund University, Vogelpothsweg 87, 44227 Dortmund, Germany, \emph{franziska.popp@tu-dortmund.de}}
\and G. Schauberger
\thanks{Chair of Epidemiology, Department of Sport and Health Sciences, Technical University of Munich, \emph{g.schauberger@tum.de}}
\and H. Van Eetvelde
\thanks{Faculty of Sciences, Department of Applied Mathematics, Computer Science and Statistics, Ghent University, Krijgslaan 281, 9000 Gent,  Belgium, \emph{hans.vaneetvelde@ugent.be}}
\and A. Zeileis
\thanks{Department of Statistics, Universit\"at Innsbruck, Austria, \emph{ Achim.Zeileis@R-project.org}}
}

\maketitle

\thispagestyle{empty}

\setlength{\parindent}{0pt}

\setlength{\columnsep}{15pt}

\textbf{Abstract}
{Three state-of-the-art statistical ranking methods for forecasting football matches are combined with 
several other predictors in a hybrid machine learning model. Namely an ability estimate for every team 
based on historic matches; an ability estimate for every team based on bookmaker consensus; average 
plus-minus player ratings based on their individual performances in their home clubs and national teams; 
and further team covariates (e.g., market value, team structure) and country-specific socio-economic factors 
(population, GDP). The proposed combined approach is used for learning the number of goals scored in the 
matches from the four previous UEFA EUROs 2004-2016 and then applied to current information to forecast the 
upcoming UEFA EURO 2020. Based on the resulting estimates, the tournament is simulated repeatedly and winning 
probabilities are obtained for all teams. A random forest model favors the current World Champion France 
with a winning probability of 14.8\% before England (13.5\%) and Spain (12.3\%). Additionally, we provide 
survival probabilities for all teams and at all tournament stages.
}\bigskip

\textbf{Keywords}:
UEFA EURO 2020, Football, Machine Learning, Team abilities, Sports tournaments.

\section{Introduction}

The use of statistical and machine learning models to predict the outcome of international football tournaments, 
such as European championships (EUROs) or FIFA World Cups,
has become pretty popular in recent years. One model class that is frequently used is the class of Poisson regression models. 
These directly model the number of goals scored by both competing teams in a single football match.
Let $X_{ij}\sim Po(\lambda_{ij})$ and $Y_{ij}\sim Po(\mu_{ij})$ denote the goals of the first and second team, respectively, in a match 
between teams $i$ and $j$, where $i,j\in\{1,\ldots,n\}$ and $n$ denotes the total number of teams 
in the regarded set of matches. For the (non-negative) intensity parameters $\lambda_{ij}$ and $\mu_{ij}$, which 
reflect the expected numbers of goals, several modeling strategies exist, which incorporate playing 
abilities or covariates of the competing teams in different ways.

In the simplest case, conditional on the teams' abilities or covariates, 
the two Poisson distributions are treated as independent.
For example, \citet{Dyte:2000} applied this model 
to data from FIFA World Cups and let the Poisson intensities of both competing teams depend 
on their FIFA ranks. \citet{GroAbe:2013} and \citet{GroSchTut:2015} considered a large set of potentially 
influential variables for UEFA EURO and World Cup data, respectively, and used $L_1$-penalized approaches 
to detect a sparse set of relevant predictors. Based on these, forecasts for the UEFA EURO~2012 
and FIFA World Cup 2014 tournaments were provided. These approaches showed that, 
when many covariates are regarded or the predictive power of the individual variables is not clear in advance,
regularized estimation approaches can be beneficial.

These approaches can be generalized in different ways to allow for dependent scores. 
For example, \citet{DixCol:97} identified a (slightly negative) correlation between the scores
and introduced an additional dependence parameter. \citet{KarNtz:2003} and \citet{GrollEtAl2018} 
modeled the scores of both teams by a bivariate Poisson distribution, which is able to account for (positive) 
dependencies between the scores. If also negative dependencies should be accounted for, 
copula-based models can be used (see, e.g., \citealp{McHaSca:2007}, \citealp{McHaSca:2011} or \citealp{boshnakov2017}).
A regularized copula regression technique was proposed by \citet{vdWurpEtAl:2020}.

Closely related to the covariate-based Poisson regression models are Poisson-based 
ranking methods for football teams. On the basis of a (typically large) set of matches, ability 
parameters reflecting the current strength of the teams can be estimated by means of maximum likelihood. 
An overview of the most frequently used Poisson-based ranking methods was provided by \citet{LeyWieEet2018}.

An alternative ranking approach that is solely based on bookmakers' odds was proposed by  
\citet{Leit:2010a}. They calculate winning probabilities for each team by aggregating winning
odds from several online bookmakers. Based on these
winning probabilities, by inverse tournament simulation
team-specific {\it bookmaker consensus abilities} can be computed by paired comparison
models, automatically stripping the effects of the tournament
draw. Next, pairwise probabilities for each
possible game at the corresponding tournament can be predicted
and, finally, the whole tournament can be simulated. 

Yet another ranking approach, the plus-minus player rating, 
calculates ratings of individual players based on the performance of their teams as a whole
using underlying match data, both on a national and international level, containing information on the starting 
line-ups as well as on certain events such as substitutions, 
red cards, and goals scored, \citep{Hv19, PaHv20}.

A fundamentally different modeling approach is based on a
random forest -- a popular ensemble learning method for classification and regression \citep{Breiman:2001}, 
which originates from the machine learning and data mining community. 
%Firstly, a multitude of so-called decision trees (\citealp{qui:1986}; \citealp{BreiFrieOls:84})
%is constructed on different training data sets, which are resampled from the original dataset. The predictions from 
%the individual trees are then aggregated, either by taking the mode of the predicted classes 
%(in classification) or by averaging the predicted values (in regression). Random forests reduce 
%the tendency of overfitting and the variance compared to regular decision trees, and are a common 
%powerful tool for prediction. 
\citet{SchauGroll2018} investigated the predictive potential of random forests in the context of international football matches and
compared different types of random forests on data containing all matches of the FIFA 
World Cups 2002--2014 with conventional regression methods for count data, such as the 
Poisson models from above. The random forests provided very satisfactory results 
and generally outperformed the regression approaches. 
\citet{GroEtAl:WM2018b} and \citet{GroEtAl:WM2019} showed on both women's and men's FIFA 
World Cup data that the predictive performance of random forests could 
be further improved by combining it with additional ranking methods, leading to what they
call a \emph{hybrid random forest model}. The term {\it hybrid} shall emphasize that
some of the features used are themselves estimates from separate statistical models.

In the present work, we carry this strategy forward and combine the 
random forest on the one hand, but also a so-called {\it extreme gradient boosting} (xgboost) approach
on the other hand, with the Poisson ranking methods from \citet{LeyWieEet2018}, the bookmaker consensus 
abilities from \citet{Leit:2010a} and plus-minus player ratings from \citet{PaHv20}.
The xgboost method is a sequential ensemble technique, which is known in the machine learning community for its high predictive power.
So in a sense, this results in {\it hybrid} or {\it combined ranking-based} machine learning approaches, similar to \citet{GroEtAl:WM2019}. 
The model is fitted to all matches of the UEFA EUROs 2004-2016 and based on the resulting estimates, 
the UEFA EURO 2020 is then simulated 100,000 times to determine winning probabilities for all 24 participating teams.

A word of caution needs to be said regarding the still ongoing COVID-19 pandemic. 
Only a very much reduced number of fans will be allowed to attend the matches and support their 
teams, the preparation of the national teams is certainly different from their usual habits 
due to sanitary restrictions, and it is not unlikely that various players will need to quarantine 
because they are tested positive at some point or have been in contact with a person affected 
by COVID-19. All these aspects, combined with the general uncertainty 
(which probably gets accentuated due to the many travels from country to country at this UEFA EURO tournament), 
are very likely to affect performances. While the impact of crowd home advantage has been studied over 
the past year for domestic competitions \citep{wunderlich2021does}, 
it is yet unknown what this implies during a competition like the UEFA European championship. 
Therefore, we do expect our predictions, as well as those of other machine learning models, 
to yield less reliable results than in normal times, since they are trained on COVID-19-free competitions.

The remainder of the manuscript is structured as follows. In Section~\ref{sec:data} we describe 
the four underlying data sets. The first data set covers all matches of the four preceding UEFA EUROs 2004-2016
including covariate information, the second consists 
of the match results of all international matches played by all national teams during certain time periods. 
The third data set contains the winning odds from several bookmakers for the single UEFA EUROs regarded in this analysis, and the
fourth is  based on match data specifying the starting line-ups together with information regarding substitutions, red cards, and goals scored.
Next, in Section~\ref{sec:methods} we briefly explain the basic idea of random forests and extreme gradient boosting, 
as well as of the three different ranking methods and, finally, how they can be combined to yield hybrid machine learning models.
In Section~\ref{sec:combine}, we fit the hybrid machine learning models to 
the data of the four UEFA EUROs 2004-2016 and investigate their predictive power. 
Based on the model fit of the model with the best predictive performance, the 
UEFA EURO 2020 is simulated 
repeatedly and winning probabilities for all teams are presented (Section~\ref{sec:prediction}). 
Finally, we conclude in Section~\ref{sec:conclusion}.

%%%%%%%%%%%%%%%%%%%%%%%%%%%%
\section{Data}\label{sec:data}

In this section, we briefly describe four fundamentally different types of data 
that can be used to model and predict international football tournaments such 
as the UEFA EURO. The first type of data covers variables that characterize 
the participating teams of the single tournaments and connects them to the results 
of the matches that were played during these tournaments. 
The second type of data is simply based on the match results of all international 
matches played by all national teams during certain time periods. These data do not only cover 
the matches from the specific tournaments but also all qualifiers and friendly matches. 
The third type of data contains the winning odds from different bookmakers separately for
single UEFA EUROs. Finally, the fourth type of data is  based on match data specifying the 
starting line-ups together with information regarding in-game events such as substitutions, red cards, and goals scored.

\subsection{Covariate data}\label{sec:covariate}

The first type of data we describe covers all
matches of the four UEFA EUROs 2004-2016 together with several
potential influence variables. Basically, we use a similar set of covariates as introduced in \citet{GrollEtAl2018}, but also added a dummy indicating whether a match is a group or a knock-out stage match. 
For each participating team, the covariates are observed either for the year of the respective UEFA EURO 
(e.g.,\ GDP per capita) or shortly before the start of the UEFA EURO (e.g.,\ average age), and, 
therefore, vary from one UEFA EURO to another.

Several of the variables contain information about the recent performance 
and sportive success of national teams, as the current form of a national team is supposed to 
have an influence on the team's success in the upcoming tournament. One additional covariate in this regard,
which we will introduce later, is reflecting the national teams' current playing abilities and 
is related to the second type of data introduced in Section~\ref{sec:historic}. 
The estimates of these ability parameters are based on a separate Poisson ranking model, 
see Section~\ref{subsec:ranking} for details, and are denoted by {\it HistAbility}. 
Another additional covariate, which is also introduced later, reflects 
the bookmaker consensus abilities on a log scale (denoted by {\it logability}) from \citet{Leit:2010a} 
and is related to the third type of data 
introduced in Section~\ref{sec:bookmaker:data}. Details on this ranking method can be found in 
Section~\ref{subsec:consensus}. The last type of additional covariates are different versions of the  
plus-minus (PM) rating of each team as proposed by \citet{PaHv20} and described in Section~\ref{sec:pm_ratings}, which
are related  to the fourth type of data introduced in Section~\ref{sec:pm:data}.

Beside these sportive variables, certain economic factors as well as variables describing 
the structure of a team's squad are collected, which are now described in more detail.

\begin{description}
\item \textbf{Economic Factors:}
\begin{description}
\item[\it GDP per capita.] 
To account for the general 
	increase of the (logarithmized) gross domestic product (GDP) during 2004--2016, a ratio of the GDP per capita of the 
	respective country and the worldwide average GDP per capita is used 
	(source: \url{http://data.worldbank.org/indicator/NY.GDP.PCAP.CD}).
%The GDP per capita represents the economic power and welfare of a nation\footnote{The GDP per capita is the gross domestic product divided by midyear population. The GDP is the sum of gross values added by all resident producers in the economy
%plus any product taxes and minus any subsidies not included in the value of the products.}. 
%Hence, countries with great prosperity might tend to focus more on sports training and promotion programs than poorer countries. 
%The GDP per capita (in US Dollar) is publicly available on the website of the World Bank 

\item[\it Population.] 
The (logarithmized) population size is used in relation to the respective global population to account for the general  world population growth\footnote{In order to collect data for all participating countries at the UEFA EUROs 2004-2020,  
different sources had to be used. Amongst the most useful ones are \texttt{http://www.wko.at}, \texttt{http://www.statista.com/} and \texttt{http://epp.eurostat.ec.europa.eu}. For some years the populations of Russia and Ukraine had 
to be searched individually.}.
%In general, larger countries have a deeper 
%pool of talented football players from which a national coach can recruit the national team squad. 
%Hence, the population size might have an influence on the playing ability of the corresponding national 
%team. However, as this potential effect might not hold in a linear relationship for arbitrarily large 
%numbers of populations and instead might diminish (see \citealp{Bern:2004}), the logarithm of the quantity is used.
\end{description}\medskip
	
\item \textbf{Home advantage:}
\begin{description}
\item[\it Host.] 
A dummy variable indicating if a national team is a hosting country. 
%There exist several studies that have analyzed the existence of a home advantage in football 
%(see, for example, \citealp{Poll:2005, Poll:2008}; \citealp{BroRaa:2002}, for FIFA World Cups 
%or \citealp{Clarke:1995}, for the English Premier league). Hence, there might also exist a home 
%effect in European championships. For this reason a dummy variable is used, indicating if a 
%national team belongs to the organizing countries.
\item[\it Neighbor.] A dummy variable indicating if a national team is from a neighboring country of the host of the UEFA EURO (including the host itself).
\end{description}\medskip

\item \textbf{Sportive factors:}
\begin{description}
%\item[\it FIFA rank.] The FIFA ranking system ranks all national 
%	teams based on their performance over the last four years (source: \url{http://de.fifa.com/worldranking/index.html}).
%\item[\it ODDSET probability.] 
%We convert bookmaker odds provided by the German state betting agency ODDSET 
%into winning probabilities (adjusting for the bookmaker's margin). 
%The variable hence reflects the probability for each team to win the respective 
%UEFA EURO\footnote{The option to bet on the European champion before the start of the 
%tournament is rather novel. ODDSET, for example, offered the bet for the first time at the UEFA EURO 2004.}.
\item[\it Market value.] 
%The market value recently has gained
%increasing attention and importance in the context of predicting the success
%of football teams (see, for example, \citealp{Gerh:2008,Gerh:2010,Gerh:2012,Gerh:2014}).
Estimates of the teams' average market values can be found on 
the webpage \texttt{http://www.transfermarkt.de}\footnote{Unfortunately,
the archive of the webpage was established not until 4th October 
2004, so the average market values of the national teams that 
we used for the UEFA EURO 2004 can only be seen as a rough approximation, 
as market values certainly changed after the UEFA EURO 2004.}.
For each national team participating in a UEFA EURO these market value estimates (on a log-scale) 
have been collected (retrospectively) right before the start of the respective tournament.
\item[\it FIFA ranking.] The FIFA ranking system ranks all national 
	teams based on their performance over the last four years\footnote{The 
	exact formula for the calculation of the underlying FIFA points and all rankings since 
	implementation of the FIFA ranking system can be found at the official FIFA 
	website: \url{http://de.fifa.com/worldranking/index.html}. Since the calculation 
	formula of the FIFA points changed after both the World Cups 2006 and 2018, the rankings according
	to FIFA points are used instead of the points. The FIFA ranking was introduced in August 1993.}. 
\item[\it UEFA points.] The associations' club coefficients rankings are 
based on the results of each association's clubs in the five previous 
UEFA CL and Europa League (previously UEFA Cup) seasons\footnote{The exact formula 
for the calculation of the underlying
UEFA points and all rankings since implementation of the UEFA ranking system can 
be found at the official UEFA website: \url{http://www.uefa.com/memberassociations/uefarankings/country/index.html}. 
The rankings determine the number of places allocated to an association (country) in the forthcoming UEFA club competitions.}.
\item[\it UEFA starting places.] The number of starting places of the corresponding
national league for both the UEFA CL and Europa League.
%Thus, the UEFA points represent the strength and success of a national league in comparison to other European national leagues. Besides, the more teams of a national league participate in the UEFA CL and the UEFA Europa League,
%the more experience the players from that national league can earn on an international level.
%As usually a relationship between the level of a national league and the level of the national team of that country is supposed,
%the UEFA points could also affect the performance of the corresponding national team. 

%\item[\it Qualification points.] Another possible assessment of the current strength of the participating national teams is their performance in the qualification matches for the respective UEFA EURO tournament. This variable contains the number of points achieved by each team in the final group standing divided by the number of matches played by each team. This qualification round usually takes place between September two years before and ends in October one year before the start of the respective UEFA EURO. The respective host teams do not have to participate and are automatically qualified. Hence, for these teams the performance from all friendly matches in the same time period was considered.

%\item[\it Qualification goals.] Similar to the {\it qualification points}, also the average goal difference from the final group standing of the preceding qualification tournament is considered.

%In August 2018 the FIFA changed from its own ranking system to the Elo ratings, which are now the basis for the FIFA World ranking list.

\end{description}\medskip

\item \textbf{Factors describing the team's structure:}

	The following variables describe the structure of the teams. 
	They were observed with the 23-player-squad\footnote{Note that due to some exceptional rules 
	related to the COVID-19 pandemic, at UEFA EURO 2020 all teams are allowed to nominate 26 players. To make those
	covariates that are based on the player numbers within the squad comparable to the historic tournaments, we multiply them by the factor 23/26.}
	nominated for the respective UEFA EUROs and were obtained manually 
	both from the website of the German football magazine {\it kicker}, \url{http://kicker.de}, and 
	from \url{http://transfermarkt.de}.%\footnote{Note that for the World Cup 2011 the size of the 
	%national teams' squads was restricted to 21 players. Hence, all of the following factors that 
	%add up players with a certain characteristic (namely all factors except for the {\it average age}) 
	%have been divided by the respective 
	%squad size (i.e.\ 21 or 23) to make them comparable across tournaments.}.
	\medskip 
	
\begin{description}
\item[\it (Second) maximum number of teammates.] For each squad, both the maximum 
	and second maximum number of teammates playing together in the same domestic team are counted.
\item[\it Absolute difference from optimal age.] First, the average age of each squad is collected. Then, the absolute distance between each team's average age and the overall average age over all teams (serving as a proxy for the optimal age) is calculated.
\item[\it Number of Champions League (Europa League) players.] 	
As a measurement of the success of the players on club level, the number of 
	players in the semi finals (taking place only few weeks before the 
	respective World Cup) of the UEFA Champions 
	League (CL) and Europa League (EL), respectively, is counted. 
\item[\it Number of players abroad/Legionnaires.] For each squad, the number of players 
	playing in clubs abroad (in the season preceding the respective UEFA EURO) is counted.
\end{description}
\end{description}

\subsubsection*{Factors describing the team's coach.}\vspace{-0.3cm}
Also attributes of a national team's coach may influence the performance of the team. Therefore,
the {\it age} of the coach (again, as absolute difference from the optimal age approximated by the overall mean) 
is observed together with a dummy variable\footnote{These 
two variables are available on several football 
data providers, see, for example, \texttt{http://www.kicker.de/}.}\saveFN\sfn, 
indicating whether the coach has the same {\it nationality} as his team or not.\medskip

\noindent In addition, we include a dummy variable indicating whether a certain match is a group- or a knockout match.
The motivation for this is that football teams might change their playing style and be more cautious in knockout matches.
In total, together with the ranking variables from the different ranking methods described in more detail below,
this adds up to 17 variables which were collected separately for each UEFA
EURO and each participating team. As an illustration, Table~\ref{data1} shows the results (\ref{tab:results}) 
and (parts of) the covariates (\ref{tab:covar}) of the respective teams, exemplarily for the first four matches 
of the UEFA EURO 2004. We use this data excerpt to illustrate how the final data set is constructed.

	\begin{table}[h]
\small
\caption{\label{data1} Exemplary table showing the results of four matches and parts of the covariates of the involved teams.}
\centering
\subfloat[Table of results \label{tab:results}]{
\begin{tabular}{lcr}
  \hline
 &  &  \\ 
  \hline
POR \includegraphics[width=0.4cm]{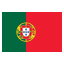} & 1-2 &  \includegraphics[width=0.4cm]{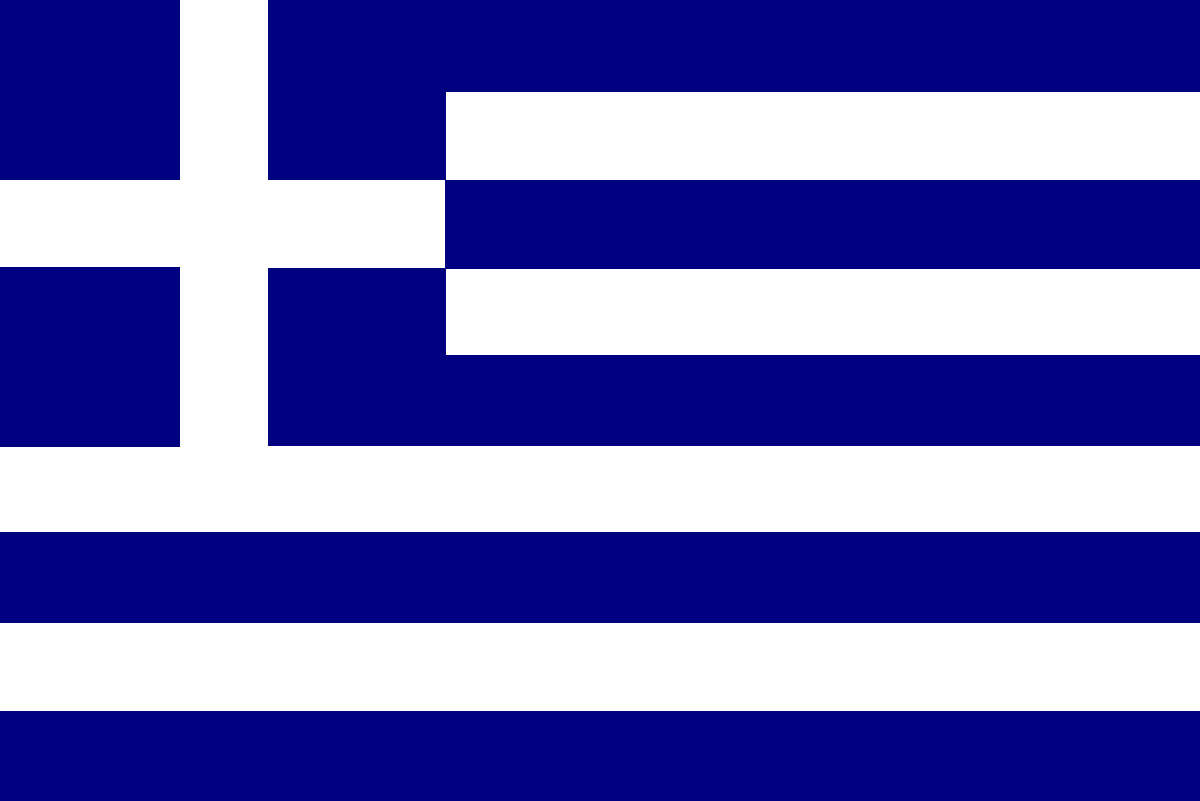} \;GRE\\
ESP\, \includegraphics[width=0.4cm]{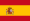} & 1-0 &  \includegraphics[width=0.4cm]{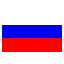} \;RUS\\
GRE \includegraphics[width=0.4cm]{GRE.png} & 1-1 &\includegraphics[width=0.4cm]{ESP.png} \;\,ESP\\
RUS \includegraphics[width=0.4cm]{RUS.png} & 0-2 &  \includegraphics[width=0.4cm]{POR.png} \;POR\\
  \vdots & \vdots & \vdots  \\
  \hline
\end{tabular}}
\hspace*{0.3cm}
\subfloat[Table of covariates \label{tab:covar}]{
\begin{tabular}{llrrrrrr}
  \hline
EURO & Team &  HistAbility & logability &  ave.PM & FIFA.rank & \ldots \\ 
  \hline
2004 & Portugal  & $1.26$  &  $0.36$ & $0.133$ & $20$ & \ldots \\ 
2004 &  Greece & $0.91$  &  $-0.38$ & $0.057$ & $34$ & \ldots\\ 
2004 &  Spain & $1.34$  & $0.33$ & $0.157$ & $3$ & \ldots \\ 
2004 &  Russia & $0.95$  & $-0.31$ & $0.076$ & $30$ & \ldots\\ 
  \vdots & \vdots & \vdots & \vdots & \vdots  &   \vdots  & $\ddots$ \\
   \hline
\end{tabular}
}
\end{table}

For the modeling techniques that we shall introduce in the following sections, all  
covariates (including the dummies for {\it Host}, {\it Continent} and {\it Nationality of coach}) 
are incorporated in the form of differences between the two competing teams. For example, the final 
variable {\it HistAbility} will be the difference between the current Poisson ability rankings of both teams. 
%The categorical variables {\it Host},
%{\it Continent} and {\it Confederation}, however, are included as separate variables for both competing teams.
%For the variable {\it Confederation}, for example, this results in two columns of the corresponding design matrix denoted by 
%{\it Confed} and {\it Confed.Oppo}, where {\it Confed} is referring to the confederation of the first-named team
%and {\it Confed.Oppo} to the one of its opponent. 

As all machine learning model introduced later use the number of goals of each team directly as the response variable, each match 
corresponds to two different observations, one per team. For the covariates, we consider 
differences which are computed from the perspective of the first-named team. 
The dummy variable {\it groupstage} corresponds to a single column in the 
design matrix and is either zero or one for both rows corresponding to the same match. For illustration, 
the resulting final data structure for the exemplary matches from Table~\ref{data1} is displayed in Table~\ref{data2}.

\begin{table}[!h]
\small
\centering
\caption{Exemplary table illustrating the data structure.}\label{data2}
\begin{tabular}{rllrrrrrr}
  \hline
Goals & Team & Opponent & Group & Historic match & Bookmaker & average PM &  FIFA rank &  ... \\ 
 &  &  & stage & abilities & abilities & player ranking &  &  ... \\ 
  \hline
     1 & Portugal & Greece & 1 & 0.34 & 0.74 & $0.076$ & $-14$ &  ...  \\ 
     2 & Greece & Portugal & 1 & $-0.34$ & $-0.74$ & $-0.076$ & 14 &  ...  \\
    1 & Spain & Russia & 1 &0.39 & 0.64 & $0.081$ & $-27$ &  ...  \\
     0 & Russia & Spain & 1 &$-0.39$ & $-0.64$ & $-0.081$ & 27 &  ...  \\ 
    1 & Greece & Spain & 1 &$-0.42$ & $-0.71$ & $-0.100$ & 31 &  ...  \\ 
    1 & Spain & Greece & 1 &0.42 & 0.71 & $0.100$ & $-31$ &  ...  \\ 
    0 & Russia & Portugal & 1 & $-0.31$ & $-0.67$ & $-0.057$ & 10 &  ...  \\
    2 & Portugal & Russia & 1 & 0.31 & 0.67 & $0.057$ & $-10$ &  ...  \\
	 \vdots & \vdots & \vdots & \vdots & \vdots & \vdots & \vdots & \vdots &  $\ddots$ \\
   \hline
\end{tabular}
\end{table}

\subsection{Historic match results}\label{sec:historic}

The data used for estimating the abilities of the teams consist of the results of every international match 
played in the last 8 years preceding the considered UEFA EURO. Besides the number of goals, 
we also need the information of the venue of the match in order to correct for the 
home effect and the moment in time when a match was played. 
 The reason is that, in the ranking method described in Section~\ref{subsec:ranking}, 
 each match is assigned a weight depending on the time elapsed since the match 
 took place. For example, Table~\ref{tab:historicdata}
 shows an excerpt of the historic match data used to obtain ability estimates for 
 the teams at the UEFA EURO 2004. 
\begin{table}[ht]
\caption{Historical results from matches prior to a tournament used for estimating 
current Poisson abilities, exemplarily for the UEFA EURO~2004}
\label{tab:historicdata}
\centering
\begin{tabular}{lllcrr}
  \hline
Date & Home team & Away team & Score & Country & Neutral \\
  \hline
2004-06-06 & Czech Republic & Estonia & 2-0 & Czech Republic & no \\
2004-06-06 & France & Ukraine & 1-0 & France & no \\
2004-06-06 & Germany & Hungary & 0-2 & Germany & no \\
2004-06-06 & Latvia & Azerbaijan & 2-2 & Latvia & no \\
\vdots & \vdots & \vdots & \vdots & \vdots & \vdots \\
   \hline
\end{tabular}
\end{table}

\subsection{Bookmaker data}\label{sec:bookmaker:data}

The basis for the bookmaker consensus model from \citet{Leit:2010a}, which is explained in more detail 
in Section~\ref{subsec:consensus}, are the ``outright'' winning odds for the entire tournament. For the upcoming UEFA EURO 2020 
these have been obtained on 2021-05-31 for all 24~teams from 19~online bookmakers via \url{https://www.oddschecker.com/} and
\url{https://www.bwin.com/}, respectively (see Table~\ref{tab:odds2020} in Appendix~\ref{sec:appendix:ata}). 
For the tournaments in 2008, 2012, and 2016 we have used the 
data from  \citet{Leit:2010a} and \citet{Zeil:2012, Zeil:2016}, respectively. For the UEFA EURO 2004 we could not find any online collections of outright winning odds but were provided with the odds from the German state betting agency ODDSET\footnote{The possibility of betting on the overall cup winner before the start of the tournament is pretty novel. The German state betting agency ODDSET offered the bet for the first time at the UEFA EURO 2020.} (upon request; see Table~\ref{tab:odds2004} in Appendix~\ref{sec:appendix:ata}). 

%The basis for the bookmaker consensus ranking model from \citet{Leit:2010a}, which is explained in more 
%detail in Section~\ref{subsec:consensus}, are the winning odds of the bookmakers\footnote{The possibility of betting on the overall cup winner before the start of the tournament is pretty novel. The German state
%betting agency ODDSET offered the bet for the first time at the UEFA EURO 2020.}. These are typically available 
%already a few weeks before the tournament start. The popularity of this specific bet has substantially increased over time:
%while for the UEFA EURO 2004 we could only obtain the odds from the German state
%betting agency ODDSET (upon request), for the subsequent UEFA EUROs we found already 
%corresponding odds from several different bookmakers publicly available.%, see Table~\ref{tab:consensus}.
%For the upcoming World Cup 2019 tournament we easily obtained the winning odds from 18 different bookmakers.

\subsection{Plus-minus player rating data}\label{sec:pm:data}

Principally, plus-minus player ratings are based on match data, both on a domestic and international level, specifying the starting 
line-ups together with information regarding certain events such as substitutions, red cards, and goals scored. For 
substitutions and red cards, additional information is required regarding which players are involved 
as well as the time when the event takes place, whereas for goals scored it suffices to know the time 
when they happen. The matches are then split into segments of maximal duration such that the set 
of players present on the pitch does not change within a segment. An example of the %corresponding
underlying data sources is displayed in Tables~\ref{tab:pmdata:1} and \ref{tab:pmdata:2}, respectively.

%Based on these data, also the information can be extracted, if important players of a national team 
%miss a UEFA EURO tournament, e.g. due to injuries.
Once this data has been used to calculate individual player ratings, it can be combined with 
information about squads to calculate different covariates: 1) The mean PM rating of the players 
in a squad, 2) the median PM rating of the players in a squad, 3) the average PM rating of the 
11 highest rated players within a squad, and 4) the number of players that were not included 
in the squad but that both had a rating that would qualify for the top 11 players in the squad and 
that had appeared in at least one match for their national team in the last two years before the tournament.

The idea of 4) is that if a team qualifies for a EURO while using their best players, and then some players are injured or otherwise missing from the final squad, then the ratings as given by e.g., Poisson-models will overestimate the quality of the team. Knowing that the squad is likely missing some key players can be helpful with respect to the predictive performance.

\begin{table}[htbp]
  \centering
  \caption{Underlying data set for deriving different plus-minus player ratings and related features, exemplarily for the UEFA EURO 2004 (part I).}
  \label{tab:pmdata:1}
    \begin{tabular}{cclll}
    \hline
    Date  & \multicolumn{1}{l}{MatchID} & Home team & Away team & Neutral \\
    \hline
    2003-08-31 & 58218 & Man. City & Arsenal & no \\
    $\vdots$ & $\vdots$ & $\vdots$ & $\vdots$ & $\vdots$ \\
    2003-09-10 & 87611 & Czech Republic & Netherlands & no \\
    $\vdots$ & $\vdots$ & $\vdots$ & $\vdots$ & $\vdots$ \\
    \hline
    \end{tabular}%
  \label{tab:addlabel}%
\end{table}%

\begin{table}[htbp]
  \centering
  \caption{Underlying data set for deriving different plus-minus player ratings and related features, exemplarily for the UEFA EURO 2004 (part II). Abbreviations used in the table: Home team (HT), away team (AT), segment ID (SID), red cards (RC), and goals scored (GS).}
  \label{tab:pmdata:2}
    \begin{tabular}{cccllcccc}
    \hline
          & & &       &       & \multicolumn{2}{l}{RC (at start)} & \multicolumn{2}{l}{GS (during)} \\
    MatchID & SID    & Time  & HT players & AT players & HT & AT & HT & AT \\
    \hline
    58218 & 1     & 0--68  & Lehmann, Tarnat, ... & Seaman, Cole, ... & 0     & 0     & 1     & 1 \\
    58218 & 2     & 68--76 & Lehmann, Tarnat, ... & Seaman, Cole, ... & 0     & 0     & 0     & 1 \\
     $\vdots$ & $\vdots$ & $\vdots$  &$\vdots$  &  $\vdots$ &  $\vdots$  & $\vdots$ & $\vdots$  & $\vdots$ \\
    87611 & 1     & 0--13  & Cech, Grygera, ... & van der Sar, Stam, ... & 0     & 0     & 0     & 0 \\
    87611 & 2     & 13--20 & Cech, Grygera, ... & van der Sar, Stam, ... & 0     & 1     & 1     & 0 \\
     $\vdots$ & $\vdots$ & $\vdots$  &$\vdots$  &  $\vdots$ &  $\vdots$  & $\vdots$ & $\vdots$  & $\vdots$ \\
    87611 & 8     & 81--94 & Cech, Ujfalusi, ... & van der Sar, Stam, ... & 0     & 1     & 1     & 0 \\
     $\vdots$ & $\vdots$ & $\vdots$  &$\vdots$  &  $\vdots$ &  $\vdots$  & $\vdots$ & $\vdots$  & $\vdots$ \\
    \hline
    \end{tabular}%
  \label{tab:addlabel}%
\end{table}%

%%%%%%%%%%%%%%%%%%%%%%%%%%%%
%%%%%%%%%%%%%%%%%%%%%%%%%%%%
\section{Hybrid machine learning models}\label{sec:methods}

In this section, we propose to use hybrid machine learning approaches that combine the 
information from all four types of data bases introduced above. 
The proposed methods combine a random forest and an extreme gradient boosting approach, respectively,
based on conventional covariate data, with the abilities estimated on the historic match results 
as used by the Poisson ranking methods, with the abilities obtained from the bookmaker consensus approach
and with the plus-minus player rankings. Before introducing the proposed hybrid method, we first separately present the 
basic ideas of the four model components.

% In this section, we briefly describe several different methods that generally come 
% into consideration when the goals scored in single football matches are 
% directly modeled. All of them (or slight modifications thereof) 
% have already been used in former research on football data and generally yielded satisfactory results.
% We aim to choose the approach that has the best performance regarding prediction and then use it to predict the FIFA World Cup 2018.

%%%%%%%%%%%%%%%%%%%%%%%%%%%%
\subsection{Random forests}\label{subsec:forest}
% \begin{itemize}
% \item Explain regression trees and forests on goals in more detail
% \item Mention that other types of forests for football data could be used as well; see Stat. Modelling
% \item But \texttt{party} regression forest performed very satisfactory and is used in the following
% \item Show plot of how tree generally works and maybe variable importance (both without abilities)
% \end{itemize}

\emph{Random forests}, originally proposed by \citet{Breiman:2001}, are an  
aggregation of a (large) number of classification or regression trees (CARTs). 
CARTs \citep{BreiFrieOls:84} repeatedly partition the predictor space mostly using 
binary splits. The goal of the partitioning process is to find partitions such that the 
respective response values are very homogeneous within a partition but very 
heterogeneous between partitions. CARTs can be used both for metric responses 
(regression trees) and for nominal or ordinal responses (classification trees). 
For prediction, all response values within a partition are aggregated either 
by averaging (in regression trees) or simply by counting and using majority vote (in classification trees).
In this work, we use trees (and, accordingly, random forests) for the prediction of 
the number of goals a team scores in a match of a UEFA EURO tournament.

Random forests are the aggregation of a large number 
$B$ (e.g., $B=5000$) of trees, grown on $B$ bootstrap samples from the original data set. 
Combining many trees has the advantage
 that the resulting predictions inherit the feature of unbiasedness from the single trees 
 while reducing the variance of the predictions. For a short introduction to random forests 
and how they can specifically 
be used for football data, see \citet{GroEtAl:WM2018b}.

% Single trees are grown independently 
% from each other. To get a final prediction, predictions of single trees are aggregated, 
% in our case of regression trees simply by averaging over all the predictions from the single trees. In order to achieve the goal that the aggregation of trees is less variant than a single tree, it is important to reduce the 
% dependencies between the trees that are aggregated to a forest. Typically, two 
% randomisation steps are applied to achieve this goal. First, the trees are not applied 
% to the original sample but to bootstrap samples or random subsamples of the data. 
% Second, at each node a (random) subset of the predictor variables is drawn which 
% are used to find the best split. These steps de-correlate the single trees and help 
% to lower the variance of a random forest compared to single trees. 
 
% The size of the 
% random subset of predictors at each node (argument \texttt{mtry}) is a tuning parameter; 
% in what follows, we will choose this parameter by cross-validation. Following 
% the suggestions of \citet{probst2017} the number of trees $B$ does not have to be tuned as long as it is chosen sufficiently large.

In \texttt{R} \citep{RDev:2018}, two slightly different variants of regression forests are 
available: the classical random forest algorithm proposed by \citet{Breiman:2001} 
from the \texttt{R}-package \texttt{ranger} \citep{ranger}, and a modification implemented 
in the function \texttt{cforest} from the \texttt{party} package\footnote{Here, the single trees are 
constructed following the principle of conditional inference trees as proposed by 
\citet{Hotetal:2006}. The main advantage of these conditional inference trees is 
that they avoid selection bias if covariates have different scales,
 e.g., numerical vs. categorical with many categories (see, for example, 
 \citealp{StrEtAl07}, and \citealp{Strobl-etal:2008}, for details). Conditional 
 forests share the feature of conditional inference trees of avoiding biased variable selection.}. 
%Cross-validation of the tuning parameter \texttt{mtry} can be done using the machine learning framework provided by the \texttt{R}-package \texttt{mlr} \citep{mlr}.
%
%Besides regression forests modeling the exact number of goals, random forests for 
%the categorical (ordinal) match outcome \textit{win}, \textit{draw} and \textit{loss} can 
%be applied. Though these forests cannot directly be used for the simulation of exact match outcomes, \citet{SchauGroll2018} explain how to suitably combine them with a random forest predicting the number of goals. Altogether, in the preliminary work from \citet{SchauGroll2018} the predictive performance of these different random forest approaches has been compared and it turned out
%that the \texttt{cforest} from the \texttt{party} package yielded the best results. For this reason, in the remainder of this work we will focus on this specific approach (from now on simply referred to as {\it Random Forest}).
In  \citet{SchauGroll2018} and \citet{GroEtAl:WM2018b}, the latter package
turned out to be superior, but both approaches are tested in this manuscript.

%%%%%%%%%%%%%%%%%%%%%%%%%%%%
\subsection{Extreme gradient boosting}\label{subsec:xgboost}

As alternative to parallel ensemble methods like the random forest approach from above,
we also consider sequential ensembles. A famous approach in this context is \emph{boosting}, 
a technique which stems from the machine learning community \citep{FreuScha:96} and was
later adapted to estimate predictors for statistical models \citep{FriHasTib:2000, Fri:2001}. 
Generally, the concept of an iterative boosting algorithm is to additively combine many weak learners 
to a powerful ensemble that achieves high accuracy \citep{Schapire:90}.
A main advantage of statistical boosting algorithms is their flexibility for high-dimensional 
data and their ability to incorporate variable selection in the fitting process \citep{MayrEtAl:pt1:2014}. 
An extensive and enlightening, general overview 
on (gradient) boosting algorithms can be found in \citet{BueHo:2008}.

\citet{Fri:2001} introduced the idea of gradient tree boosting, using decision trees as learners.
The decision trees are repeatedly fitted on the residuals of the previous fit and, hence, 
are combined to a sequential ensemble.
This technique was then further improved by \citet{chen2016xgboost} via introducing additional 
regularization in the objective function. %In particular, on the one hand specific regularization terms make the single trees weak learners, on the other hand in a certain boosting iteration the next tree is additively incorporated into the ensemble via a weak learning rate. 
The regularization terms make the single trees weak learners to avoid overfitting. In a certain boosting iteration, the next tree is additively incorporated into the ensemble after multiplication with a rather small learning rate which makes the learners even weaker.
The method is called \emph{extreme gradient boosting}, in short \emph{xgboost},
and is known in the machine learning community for its high predictive power. It has
been very successful in prestigious machine learning competitions,
such as those organized by Kaggle (\url{https://www.kaggle.com}).
The approach is implemented in the \texttt{xgb.train} function from the \texttt{xgboost} R package \citep{chen:2021}.

One important aspect is that xgboost involves several tuning parameters, such as e.g.\ the 
learning rate, the optimal number of boosting steps and several penalty parameters. 
For this purpose, we specified reasonable, discrete parameter grids
and used multivariate 10-fold cross validation via the \texttt{xgb.cv} function
to determine optimal tuning parameters.
%\red{For technical details of the xgboost method, see Appendix~\ref{sec:xgboost:tec}.}

%%%%%%%%%%%%%%%%%%%%%%%%%%%%
\subsection{Current ability ranking based on historic matches}\label{subsec:ranking}

In this section we describe how (based on historic match data, see Section~\ref{sec:historic}) Poisson models can be used to obtain rankings that reflect a team's current ability. We will restrict our attention to the best-performing model according to the comparison achieved in \cite{LeyWieEet2018}, namely the bivariate Poisson model. The main idea consists in assigning a strength parameter to every team and in estimating those parameters over a   period of $M$ matches via weighted maximum likelihood based on time depreciation.

The time decay function is defined as follows: a match  played $x_m$ days back gets a weight of 
\begin{equation*}\label{smoother}
w_{time,m}(x_m) = \left(\frac{1}{2}\right)^{\frac{x_m}{\mbox{\small Half period}}},
\end{equation*}
meaning that, for instance, a match played \emph{Half period} days ago only contributes half as much as a match played today. We stress that the \emph{Half period} refers to calendar days in a year, not match days. In the present case we use a Half period of 3 years (i.e. 1095 days) based on an optimization procedure to determine which Half period led to the best prediction for men's football matches based on the average Rank Probability Score (RPS; \citealp{gnei:2007})

The bivariate Poisson ranking model is based on a proposal from \cite{KarNtz:2003} and can be described as follows. If we have $M$ matches featuring a total of $n$ teams, we write $Y_{ijm}$  the random variable \textit{number of goals scored by team $i$ against team $j$ ($i,j\in \{1,...,n\}$) in match $m$} (where $m \in \{1,...,M\}$). The joint probability function of the home and away score is then given by the bivariate Poisson probability mass function, 
%\begin{eqnarray*} 
$$
{\rm P}(Y_{ijm}=z, Y_{jim}=y) = %&=&\nonumber\\
%&& 
\frac{\lambda_{ijm}^z \lambda_{jim}^y}{z!y!} \exp(-(\lambda_{ijm}+\lambda_{jim}+\lambda_{C}))%\cdot\,
\sum_{k=0}^{\min(z,y)} \binom{z}{k} \binom{y}{k}k!\left(\frac{\lambda_{C}}{\lambda_{ijm}\lambda_{jim}}\right)^k, 
%\end{eqnarray*}
$$
where $\lambda_{C}\geq0$ is a covariance parameter assumed to be constant over all matches and $\lambda_{ijm}\geq0$ is the expected number of goals for team $i$ against team $j$ in  match $m$, which we model as
\begin{eqnarray}
\label{independentpoisson}\log(\lambda_{ijm})&=&\beta_0 + (r_{i}-r_{j})+h\cdot \mathds{1}(\mbox{team $i$ playing at home})\,,
\end{eqnarray}
where $\beta_0\in\mathds{R}$ is a common intercept and $r_i, r_j \in\mathds{R}$ are the strength parameters of team~$i$ and team~$j$, respectively. Since the ratings are unique up to addition by a constant, we add the constraint that the sum of the ratings has to equal zero. The last term $h\in\mathds{R}$ represents the home effect and is only added if team~$i$ plays at home. We get an independent Poisson model if $\lambda_C=0$. The overall  (weighted) likelihood function then reads
\begin{equation*}
L = \prod_{m=1}^{M}\left({\rm P}(Y_{ijm}=y_{ijm}, Y_{jim}={y_{jim}})\right)^{w_{time,m}}, 
\end{equation*}
where $y_{ijm}$ and $y_{jim}$ stand for the actual number of goals scored by teams $i$ and $j$ in match $m$. The values of the strength parameters $r_1,\ldots,r_n$, which allow ranking the different teams, are computed numerically as maximum likelihood estimates {on the basis of historic match data as described in Section~\ref{sec:historic}}.  These parameters also allow to predict future match outcomes thanks to the Equation~\eqref{independentpoisson}.

\subsection{Bookmaker consensus model}\label{subsec:consensus}

Prior to the tournament, on 2021-05-31, we obtained long-term winning odds from 19~online bookmakers.
However, before these odds can be transformed to winning probabilities, the stake has to be accounted for
and the profit margin of the bookmaker (better known as the ``overround'') has to be removed
\citep[for further details see][]{ref:Henery:1999, ref:Forrest+Goddard+Simmons:2005}. 
Here, it is assumed that the quoted odds are derived from the underlying ``true'' odds as:
$\mbox{\it quoted odds} = \mbox{\it odds} \cdot \delta + 1$,
where $+ 1$ is the stake (which is to be paid back to the bookmakers' customers in case they win)
and $\delta < 1$ is the proportion of the bets that is actually paid out by the bookmakers.
The overround is the remaining proportion $1 - \delta$ and the main basis of the bookmakers'
profits (see also \citealp{ref:Wikipedia:2019} and the links therein). 
Assuming that each bookmaker's $\delta$ is constant across the various teams in the tournament
\citep[see][for all details]{Leit:2010a}, we obtain overrounds for all
bookmakers with a median value of 17.3\%.

To aggregate the overround-adjusted odds across the 19~bookmakers, we transform them to
the log-odds (or logit) scale for averaging \citep[as in][]{Leit:2010a}.
The bookmaker consensus is computed as the mean winning log-odds for each team across bookmakers
and then transformed back to the winning probability scale.

In a second step the bookmakers' odds are employed to infer the contenders' relative abilities
(or strengths). To do so, an ``inverse'' tournament simulation based on team-specific
abilities is used. The idea is the following:
\begin{enumerate}
  \item If team abilities are available, pairwise winning probabilities can be derived
    for each possible match using the classical \cite{ref:Bradley+Terry:1952} model. This model is
    similar to the Elo rating \citep{ref:Elo:2008}, popular in sports, and computes the
    probability that a Team~$A$ beats a Team~$B$ by their associated abilities (or strengths): 
     \[
    \mathrm{Pr}(A \mbox{ beats } B) = \frac{\mathit{ability}_A}{\mathit{ability}_A + \mathit{ability}_B}.
    \]
  \item Given these pairwise winning probabilities, the whole tournament can be easily
    simulated to see which team proceeds to which stage in the tournament and which
    team finally wins\footnote{By 
    adopting the classical Bradley-Terry model, the simulation of each
  match yields only a winner and a loser, without the possibility of a tie
  or any further information about the number of goals or the goal
  difference. This is sufficient for the knock-out stage of the
  tournament, as it reflects the fact that the actual matches always have
  a winner (if necessary through overtime and penalties). However, for the
  group phase within the simulation this approach might result in tied
  teams. If necessary, such ties are resolved through additional
  ``fictitious'' matches between the tied teams in order to obtain unique
  winners and runner-ups of the groups.}.
  \item Such a tournament simulation can then be run sufficiently often (here 100,000 times)
    to obtain relative frequencies for each team to win the tournament.
\end{enumerate}
Here, we use the iterative approach of \cite{Leit:2010a}
to find team abilities so that the resulting simulated winning probabilities
(from 100,000 runs) closely match the bookmaker consensus probabilities. This allows to
strip the effects of the tournament draw (with weaker/easier and stronger/more difficult
groups), yielding a log-ability measure (on the log-odds scale) for each team.

\subsection{Plus-minus player ratings}
\label{sec:pm_ratings}

This section describes a method for calculating ratings of individual players based on the performance of their teams as a whole, 
known as \textit{plus-minus (PM) ratings} \citep{Hv19}. The starting point is match data specifying the starting line-ups together with information regarding substitutions, red cards, and goals scored, see Section~\ref{sec:pm:data} for more details. 

The basic idea of adjusted PM ratings is to formulate a regression model where the dependent variable corresponds to the observed goal difference within a segment and the covariates include indicator variables for the presence of players. The regression coefficients of the player specific indicator variables can then be interpreted as player ratings. In a basic form, let $y_i$ be the observed goal difference for segment $i$, taken from the perspective of the home team. Furthermore, let $x_{ij} = 1$ if player $j$ appears on the pitch for the home team during segment $i$, $x_{ij} = -1$ if the player appears for the away team, and $x_{ij} = 0$ otherwise. Then, player ratings $\beta_j$ can be obtained as the estimated regression coefficients of a simple linear regression:
$$
y_i = \sum_{j} \beta_j x_{ij} + \epsilon_i,
$$
where $\epsilon_i$ is an error term. When applying PM ratings to football players, the above specification is too simplistic, and additional covariates and estimation tricks must be incorporated. We use here the ratings as described by \citet{PaHv20}, including the following adjustments:

\begin{itemize}
\item A country-specific home-field advantage is modelled by additional covariates.
\item Red cards are handled by introducing additional covariates for each potential 
red card and by weighing the appearances of the remaining players.
\item Players' ratings are adjusted by a factor that depends on their age.
\item Another adjustment of players' ratings is made based on the set of league tournaments in which they have participated.
\item Each observation is weighted based on three criteria: the time since the match was played, 
the duration of the corresponding segment, and the goal difference at the beginning and end of the segment.
\item The estimation of ratings is based on regularization in the form of ridge regression (see \citealp{HoeKen:70}).
\item For coefficients corresponding to player ratings, the regularization is adjusted ad hoc, under 
the assumption that a player is more likely to be of similar playing strength as the most common teammates, rather than an average player.
\end{itemize}

The resulting PM ratings have been examined in past studies. \citet{GeHv20} showed how the 
ratings are related to certain key-performance indicators based on event data, while \citet{HvGe21} 
compared them to an alternative rating system based on valuing individual player actions. \citet{ArHv20} 
found that the PM ratings of players in the starting line-ups of teams contained relevant information
for predicting match outcomes, in particular in combination with a team rating based on the Elo system. 

For this project, we calculated the average, median and ``best-11-player'' PM team rankings as well 
as the number of ``missing important PM players'' as introduced in Section~\ref{sec:pm:data}. 
The first three variables are extremely correlated ($>0.98$). During the tuning of 
our models we found that the {\it average PM player rating} was slightly outperforming the other
two, so we only included it together with the missing player in the hybrid machine learning models,
which we introduce in the next section.% and which are compared with regard to predictive power in Section~\ref{sec:combine}

%For the modelling and prediction of UEFA EURO tournaments, we consider four factors derived from 
%the PM ratings: 1) The mean PM rating of the players in a squad, 2) the 
%median PM rating of the players in a squad, 3) the average PM rating of the 11 highest 
%rated players within a squad, and 4) the number of players that were not included in the squad 
%but that both had a rating that would qualify for the top 11 players in the squad and that had 
%appeared in at least one match for their national team in the last two years before the tournament.

\subsection{Combine methods to hybrid machine learning models}

In order to link the information provided by the covariate data, the historic match data, the bookmakers' odds
and the plus-minus player rating related data, we now combine the random forest approach from 
Section~\ref{subsec:forest} and the extreme gradient boosting approach from Section~\ref{subsec:xgboost}
with the ranking methods from Sections~\ref{subsec:ranking}-\ref{sec:pm_ratings}. 
We propose to use the ranking approaches to generate new (highly informative) covariates 
that can be incorporated into the statistical models. 
For that purpose, we estimate current Poisson ranking team abilities $r_i$ based 
on historic match data (see Section~\ref{subsec:ranking}) as well as 
for the ``\# of missing players" (see Section~\ref{sec:pm_ratings}). 

Moreover, for each UEFA EURO we use the winning odds provided by the bookmakers and
calculate the team log-abilities $s_i, i=1,\ldots,N$, of all $N\in\{16,24\}$ 
participating teams shortly before the start of the respective tournament (see Section~\ref{subsec:consensus}).
This procedure gives us the estimates $\hat s_i$ as an additional covariate covering the %current 
strength for all teams participating in a certain UEFA EURO. Actually, this variable 
turns out to be much more informative than e.g.\ the FIFA ranking, see Section~\ref{sec:fitforest}. 

 Finally, based on historic match segment data, 
we estimate the (average) plus-minus team rankings $pm_i, i=1,\ldots,N$, of all 
participating teams shortly before the start of the respective tournament (see again Section~\ref{sec:pm_ratings}). 
The corresponding estimates $\widehat{pm}_i$ again serve as another additional covariate. 
Also this variable turns out to be 
 rather relevant, see again Section~\ref{sec:fitforest}.

The newly generated variables can be added to the covariate data based on 
previous UEFA EUROs and a random forest, an xgboost model or actually any 
other statistical or machine learningmodel, such as
 e.g.\ a lasso-regularized regression model\footnote{The idea of Lasso penalization was 
 first introduced by \citet{Tibshirani:96}. %(see Appendix~\ref{sec:lasso}), 
Such a Lasso regression model can be easily tuned and fitted using the function \texttt{cv.glmnet} from the \texttt{R}-package \texttt{glmnet} \citep{FrieEtAl:2010}.}, can be fitted to these data.  
Lasso regression was used for example in \citet{GroSchTut:2015} to predict the FIFA World Cup 2014.
More details on how classical regression approaches can be used
for the modeling and prediction of football matches can also be found in \citet{GroSchauEet:2020}. 
Based on these models, new matches (e.g., matches from an upcoming UEFA EURO tournament) can be predicted.
Exemplarily for the random forest, a new observation is predicted 
by dropping down its covariate values from each of the $B$ 
regression trees, resulting in $B$ distinct predictions. The average of those is then used as a 
point estimate of the expected numbers of goals conditioning on the covariate values. 
Similarly, also for the xgboost and the Poisson Lasso regression model the covariate values
of the new observation can be plugged into the sequential tree ensemble and the corresponding linear predictor, respectively.
In order to be able to use the point estimates from the tree-based models for the prediction of the 
outcome of single matches or a whole tournament, we follow \citet{GroEtAl:WM2018b} 
and treat the predicted expected value for the number of goals as 
an estimate for the intensity $\lambda$ of a Poisson distribution $Po(\lambda)$.
For the Poisson lasso regression model, this is implicitly done anyway.
%However, these point estimates cannot directly be used for the prediction of the 
%outcome of single matches or a whole tournament. First of all, plugging in both predictions 
%corresponding to one match does not necessarily deliver an integer outcome (i.e., a result). 
%For example, one might get predictions of 2.3 goals for the first and 1.1 goals for the second team. 
%Furthermore, as no explicit distribution is assumed for these predictions it is not 
%possible to randomly draw results for the respective match. Hence, following \citet{GroEtAl:WM2018b} 
%we treat the predicted expected value for the number of goals as an estimate for the intensity $\lambda$ of a Poisson 
%distribution $Po(\lambda)$. This procedure could be motivated by assuming 
%that within each terminal node of a tree we fit a simple intercept-only Poisson model 
%where the average of all scores equals the maximum likelihood estimate of the intercept parameter. 
This way we can randomly draw results for single matches and compute 
probabilities for the match outcomes \textit{win}, \textit{draw} and \textit{loss} by 
using two independent Poisson distributions (conditional on the covariates) for both scores.

%%%%%%%%%%%%%%%%%%%%%%%%%%%%
\section{Model performance}\label{sec:combine}

%%%%%%%%%%%%%%%%%%%%%%%%%%%%

% \begin{itemize}
% \item Maybe show here ``old" tables from Stat. Modelling (i.e. without abilities as covariate) together with ranking model performance
% \item Explain why (they use/are based on different information) it might lead to improvement to combine the approaches, meaning to use abilities as a covariate in the RF and regression models
% \item Explain how they are combined
% \item Then show in ``new" tables how the models have improved
% \item The RFs still perform best for most criteria and for the quadratic errors for goals and goal difference at least come close to the ranking methods; for the rest of the paper, hence, the RF combined with the ranking abilities as covariate will be used
% \end{itemize}

In the following, we investigate the predictive performance of the proposed hybrid machine learning models
and compare them also to a more conventional (regularized) Poisson regression approach. 
% On the one hand, we compare the hybrid model to its separate components, 
% namely the pure random forest (without additional team abilities)
% and to the ranking method. On the other hand, we use a Lasso 
% regression approach as an additional reference method. 
This method is also ``hybrid'' in the sense that it includes both the covariate data from Section~\ref{sec:covariate} and 
the ranking variables from Sections~\ref{subsec:ranking}-\ref{sec:pm_ratings}. It links the feature 
information to the number of goals in a log-linear Poisson model. The model is estimated 
using a penalized likelihood approach. The details for this method can also be found in \citet{GroSchauEet:2020}.
%Appendix~\ref{sec:lasso}. 
%Analogous to the proposed random forest method, the Lasso regression method can also be extended by incorporating the team abilities from the ranking method as a further covariate. Accordingly, these two methods are from now on referred to as \emph{Lasso} and \emph{Hybrid Lasso}, respectively.

Altogether, the following four approaches are now compared with regard to their predictive performance:
two random forest implementations, \texttt{ranger} and \texttt{cforest}, the \emph{xgboost} method and
a conventional \emph{lasso} Poisson regression model. 
For this purpose, we  apply the following general procedure on the UEFA EURO 2004-2016 
data for all methods: %except for the ranking approach:
\begin{enumerate}{\it
\item Form a training data set containing three out of four UEFA EUROs.\vspace{0.1cm}
\item Fit each of the methods to the training data.\vspace{0.1cm}
\item Predict the left-out UEFA EURO using each of the prediction methods.\vspace{0.1cm}
\item Iterate steps 1-3 such that each UEFA EURO is once the left-out one.\vspace{0.1cm}
\item Compare predicted and real outcomes for all prediction methods.}\vspace{-0.1cm}
\end{enumerate}
This procedure ensures that each match from the total data set is once part of the 
test data and we obtain out-of-sample predictions for all matches. 
%For the ranking model, we do not have to apply the described iterative procedure 
%and can directly jump to step~{\it 5}. Instead, the model was fit to a large data set 
%covering all historic matches from the past 8 years up to the start of the respective 
%World Cup that shall be predicted. The corresponding ability parameter estimates 
%can then be used directly for the prediction.
In step~{\it 5}, several different performance measures for the quality of the predictions are investigated.

Let $\tilde y_i\in\{1,2,3\}$ be the true ordinal match outcomes for all $i=1,\ldots,N$ 
matches  from the four considered UEFA EUROs. Additionally, let $\hat\pi_{1i},\hat\pi_{2i},\hat\pi_{3i},~i=1,\ldots,N$, 
be the predicted probabilities for the match outcomes obtained by one of the different methods mentioned above. 
These can be computed by assuming that the numbers of goals follow (conditionally) 
independent Poisson distributions, where the event rates $\lambda_{1i}$ and $\lambda_{2i}$ 
for the scores of match $i$ are estimated by the respective predicted expected values. Let 
$G_{1i}$ and $G_{2i}$ denote the random variables representing the number of goals 
scored by two competing teams in match $i$. Then, the probabilities $\hat \pi_{1i}=P(G_{1i}>G_{2i}), 
\hat \pi_{2i}=P(G_{1i}=G_{2i})$ and $\hat \pi_{3i}=P(G_{1i}<G_{2i})$, which are based on the 
corresponding Poisson distributions $G_{1i}\sim Po(\hat\lambda_{1i})$ and $G_{2i}\sim Po(\hat\lambda_{2i})$ 
with estimates $\hat\lambda_{1i}$ and $\hat\lambda_{2i}$, can be easily calculated via the 
Skellam distribution. For a short description of the Skellam distribution, see Appendix~\ref{sec:general}.
Based on these predicted probabilities, we use three different performance measures 
to compare the predictive power of the methods:
\begin{itemize}
\item the multinomial {\it likelihood}, which for a single match outcome is defined as $\hat \pi_{1i}^{\delta_{1\tilde y_i}} \hat \pi_{2i}^{\delta_{2\tilde y_i}} \hat \pi_{3i}^{\delta_{3 \tilde y_i}}$, with $\delta_{r\tilde y_i}$ denoting Kronecker's delta, which is defined in Appendix~\ref{sec:general}. The multinomial likelihood reflects the probability of a correct prediction. Hence, a large value reflects a good fit.\vspace{0.1cm}
\item  the {\it classification rate}, based on the indicator functions $\mathds{1}(\tilde y_i=\underset{r\in\{1,2,3\}}{\mbox{arg\,max }}(\hat\pi_{ri}))$, indicating whether match $i$ was correctly classified.
Again, a large value of the classification rate reflects a good fit.
However, note that along the lines of \citet{gnei:2007} the  classification rate does not constitute a proper scoring rule.\vspace{0.1cm}
\item the {\it rank probability score} (RPS), which, in contrast to both measures introduced above, explicitly accounts for the ordinal structure of the responses. 
For our purpose, it can be defined as $\frac{1}{3-1} \sum\limits_{r=1}^{3-1}\left( \sum\limits_{l=1}^{r}(\hat\pi_{li} - \delta_{l\tilde y_i})\right)^{2}$. As the RPS is an error measure, here a low value represents a good fit.
\end{itemize}
Odds provided by bookmakers serve as a natural benchmark for these predictive 
performance measures. For this purpose, we collected the so-called ``three-way'' odds 
for (almost) all matches of the UEFA EUROs 2004-2016\footnote{Three-way odds 
consider only the match tendency with possible results \emph{victory team 1}, \emph{draw} 
or \emph{defeat team 1} and are usually fixed some days before the corresponding match 
takes place. This allows the bookmakers to incorporate current information (e.g., injuries of 
important players) into the odds during the run of a tournament. The three-way odds were obtained 
from the website \url{http://www.betexplorer.com/}.}. 
%Unfortunately, for 6 matches from the FIFA World Cup 2006 no odds were 
%available, hence, the results from Table~\ref{tab:probs} are based on 250 matches only.}.
By taking the three quantities $\tilde \pi_{ri}=1/\mbox{odds}_{ri}, r\in\{1,2,3\}$, of a 
match $i$ and by normalizing with $c_i:=\sum_{r=1}^{3}\tilde \pi_{ri}$ in order to adjust 
for the bookmaker's margins, the odds can be directly transformed into probabilities using $\hat \pi_{ri}=\tilde \pi_{ri}/c_i$
\footnote{The transformed probabilities implicitly assume that the bookmaker's margins are equally distributed on the three possible match tendencies.}.
% Using these predicted probabilities $\hat \pi_{ri}$, we can evaluate the three performance measures for (ordinal) match outcomes introduced above also for the information contained in bookmakers' odds.

Table~\ref{tab:probs} displays the results for these (ordinal) performance measures 
for the four prediction methods as well as for the bookmakers, 
averaged over 144 matches from the four UEFA EUROs 2004-2016 (regarding the results in regular time, i.e. after 90 minutes and without possible extra time or penalty shootout).
First of all, it turns out that all approaches achieve a slightly worse performance compared to our previous analysis on
FIFA World Cups in \citet{GroEtAl:WM2018b}. 

The hybrid cforest slightly outperforms the other approaches 
with respect to the multinomial {\it likelihood}, closely followed by the xgboost approach, and is also (together with xgboost)
 the best regarding the {\it classification rate}, getting even close to the bookmakers as the natural benchmark. 
In terms of the RPS, all four methods are rather comparable with a slight advantage for the lasso Poisson regression model. 
%while in terms of the classification rate, the cforest and xgboost . 

%While the average likelihood is only slightly lower than for the bookmakers, the RPS of the hybrid 
%random forest is the best among all competitors. In general, the random forest approaches also stand 
%out for their high classification rates, only being slightly outperformed by the hybrid Lasso. The two Lasso methods 
%yield satisfactory results with respect to most criteria, only in terms of RPS they are outperformed by the other 
%methods. Furthermore, the hybrid Lasso approach also shows that the incorporation of the ability estimates from 
%the ranking method can provide valuable additional information compared to the remaining covariates.

\begin{table}[H]
\small
\caption{\label{tab:probs}Comparison of the prediction methods for ordinal match outcomes 
\emph{victory team 1}, \emph{draw} or \emph{defeat team 1} (regarding the results in regular time, i.e. after 90 minutes and without possible extra time or penalty shootout; best performing approach in bold font); additionally, 
the predictions based on the bookmakers' odds are shown as a natural benchmark (bottom line).}\vspace{0.2cm}
\centering
\begin{tabular}{lrrr}
  & Likelihood & Class. Rate & RPS \\ 
  \toprule
  ranger & 0.372 & 0.458 & 0.216 \\ 
%\midrule
  cforest & {\bf 0.382} & {\bf 0.486} & 0.213 \\ 
%\midrule
  xgboost & 0.380 & {\bf 0.486} & 0.217 \\ 
%\midrule
lasso & 0.379 & 0.458 & {\bf 0.210} \\ 
\midrule
  bookmakers & 0.400 & 0.493 & 0.203 \\ 
\bottomrule
\end{tabular}
\end{table}

As the proposed methods can also be used to simulate the tournament course of an upcoming tournament 
(see also Section~\ref{sec:simul} for a prediction of the UEFA EURO 2020),
we are also interested in the performance of the regarded methods with respect to the prediction of the 
exact number of goals. In order to identify the teams that qualify for the knockout stage, the precise final group 
standings need to be determined. To be able to do so, the precise results of the matches in the group 
stage play a crucial role\footnote{The final group standings are determined by (1) the number of
points, (2) the goal difference and (3) the number of scored goals.
If several teams coincide with respect to all of these three criteria, a
separate chart is calculated based on the matches between the coinciding
teams only. Here, again the final standing of the teams is
determined following criteria (1)--(3). If still no distinct decision can
be taken, the decision is induced by lot.\label{fifa:rules}}.

For this reason, we also evaluate the methods' performances
with regard to the quadratic error between the observed and predicted
number of goals for each match and each team, as well as between the observed and predicted goal 
difference. Now let $y_{ijk}$, for $i,j=1,\ldots,n$ and $k\in\{2004,2008,2012,2016\}$,
denote the observed numbers of goals scored by team $i$ against team $j$ in tournament $k$ and
$\hat y_{ijk}$ a corresponding predicted value, obtained by one of the compared methods. 
Then we calculate the two absolute errors $|y_{ijk}-\hat y_{ijk}|$ and 
$\left|(y_{ijk}-y_{jik})-(\hat y_{ijk}-\hat y_{jik})\right|$ for all $N$ matches of the four 
UEFA EUROs 2004-2016. Finally, per method we calculate (mean) absolute errors. Note that in this case 
the odds provided by the bookmakers cannot be used for comparison. 
Table~\ref{tab:goals} shows that actually all four methods yield rather similar results,
with slight advantages for lasso.%, but again closely followed by the cforest and xgboost approach.
\begin{table}[H]
\small
\caption{\label{tab:goals}Comparison of the prediction methods for the exact number of goals
and the goal difference based on mean absolute error (best performing approach in bold font).}\vspace{0.2cm}
\centering
\begin{tabular}{lrr}
  & Goals & Goal Difference \\ 
  \toprule
  ranger & 0.862 & 1.176 \\ 
%\midrule
  cforest & 0.862 & 1.166 \\ 
%\midrule
  xgboost & 0.883 & 1.162 \\ 
%\midrule
lasso & {\bf 0.846} & {\bf 1.148} \\ 
   \bottomrule
\end{tabular}
\end{table}

Altogether, based on these results and our positive experiences in earlier research projects \citep{GroEtAl:WM2018b, GroEtAl:WM2019}, 
we assess the hybrid cforest method to be the best and most reliable choice for forecasting
the upcoming UEFA EURO 2020 tournament. 

But note that also the xgboost approach seems to be principally very promising for modeling and predicting football matches. 
However, we experienced some higher sensitivity and instability during the (more sophisticated) tuning process compared to the other methods, which we believe is mostly due to small sample size of our training data. In particular, as we had to
always exclude one full UEFA EURO from the training data in our leave-one-tournament-out strategy in order
 to assess the models' performances on external validation data, we believe that the full potential of xgboost is not yet fully exploited in this competition.

%%%%%%%%%%%%%%%%%%%%%%%%%%%%
\section{Modeling the UEFA EURO 2020}\label{sec:prediction}
We now fit the proposed hybrid cforest model to the full UEFA EURO 2004-2016 data. 
Next, we calculate the Poisson ranking ability parameters 
based on historic match data over the 8 years preceding the UEFA EURO 2020, 
as well as the bookmaker consensus abilities based on
the winning odds from 19 different bookmakers, and the average PM player ratings 
as well as the number of important PM players missing in the squad.
Based on conventional covariate data and those additional ability and rating variables, the fitted cforest model will be used to 
simulate the UEFA EURO 2020 tournament 100,000 times to determine winning probabilities for all 24 participating teams.

\subsection{Fitting the hybrid cforest to the UEFA EUROs 2004-2016 data}\label{sec:fitforest}

\begin{figure}[!ht]
	\centering
		\includegraphics[width=1.1\textwidth]{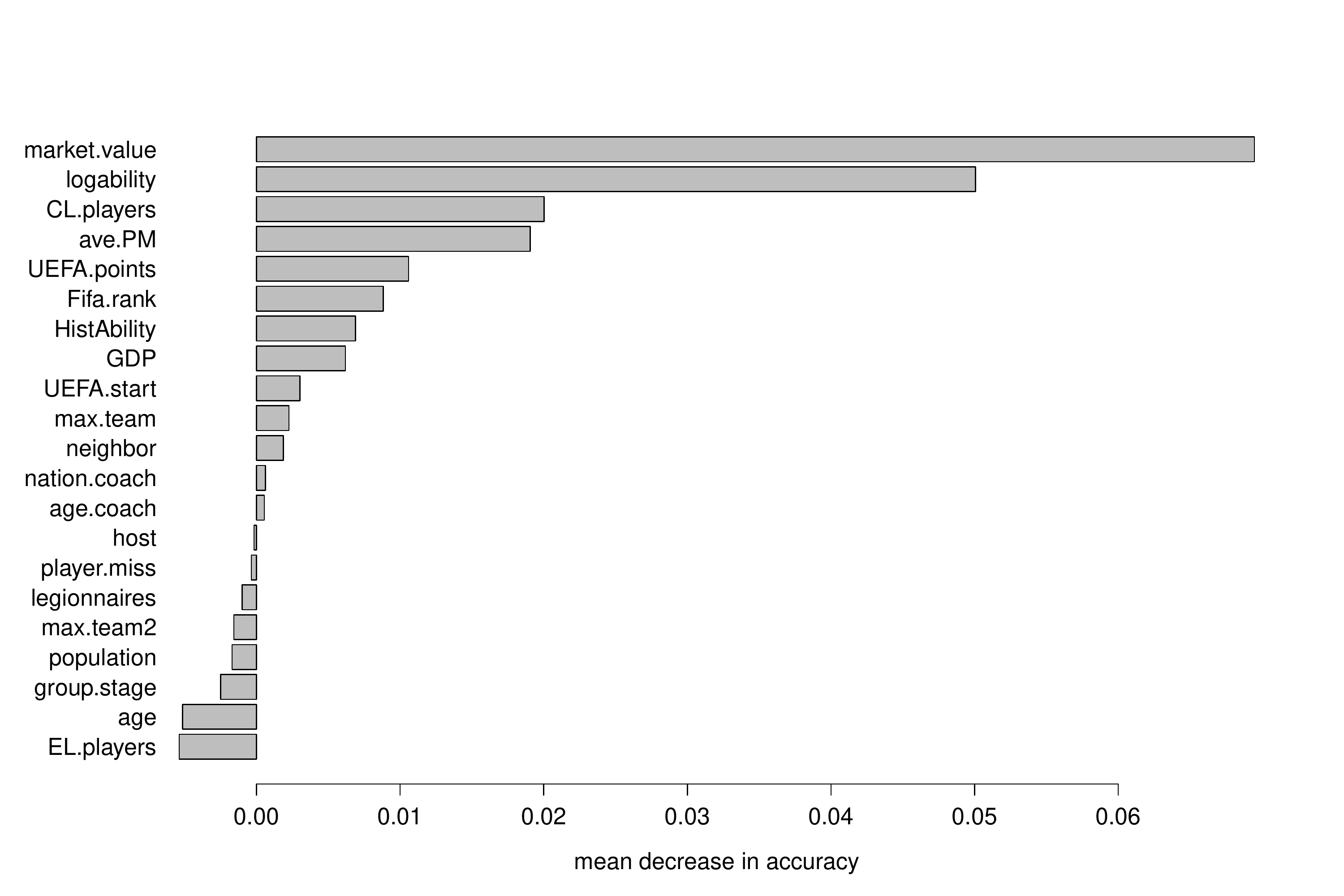}
	\caption{Bar plot displaying the variable importance in the hybrid cforest model applied to UEFA EURO 2004-2016 data.}
	\label{var_imp}
\end{figure}

We next fit the hybrid cforest to the 
complete data set covering the four UEFA EUROs 2004-2016. 
%Tuning is done by the help of the \texttt{xgb.cv} function from the \texttt{xgboost} package.
As suggested in regression settings, the optimal number of input variables randomly sampled as candidates 
at each node is set to  \texttt{mtry}$=\lc\sqrt{p}\rc=4$.
%Of course, it would be appealing to visualize and interpret the obtained 
%results in order to learn about the relationship
%between the sporting success of a men's national soccer team and the set of possible 
%influence variables. However, in contrast to regression trees, random forests are much harder to visualize 
%and to interpret. While for individual trees the effect of a single predictor can be derived from 
%the respective dendrogram graph, this is almost impossible for random forests.
%Each predictor may have different effects (or no effect at all) in different trees. 
The  best way to understand the role of the single predictor variables in a complicated, blackbox-type 
machine learning model such as the cforest is the so-called variable importance \citep{Breiman:2001}.
Typically, the variable importance of a predictor is measured by permuting each of the predictors 
separately in the out-of-bag observations of each tree. Out-of-bag observations 
are observations which are not part of the respective subsample or bootstrap 
sample that is used to fit a tree. Permuting a variable means that within the 
variable each value is randomly assigned to a location within the vector. 
If, for example, \emph{FIFA.rank} is permuted, the FIFA.rank of the German team
 in 2004 could be assigned to the FIFA.rank of the Spanish team in 2016. 
 When permuting variables randomly, they lose their information with respect 
 to the response variable (if they have any). Then, one measures the loss of prediction 
 accuracy compared to the case where the variable is not permuted. 
 Permuting variables with a high importance will lead to a higher loss of 
 prediction accuracy than permuting values with low importance.
%To illustrate the concept of variable importance, 
Figure~\ref{var_imp} shows bar plots of the variable importance values for all variables in the 
hybrid cforest applied to the data of the UEFA EUROs 2004-2016. 
Interestingly, the market value is the most important predictor
in the cforest model, followed by  %and carries clearly more information than all other predictors. But also 
the abilities from the bookmaker consensus approach. But also the {\it number of CL players}, the {\it average PM player rating}
and the {\it UEFA points} seem to be more informative compared e.g.\ to the \emph{FIFA rank}.
Besides those variables, also the \emph{Poisson ranking abilities} and the {\it GDP}
contain relevant information concerning the current strengths of the teams. 
Hence, it is definitely worth the effort to estimate such abilities in separate statistical models. 
For a more detailed comparison of the team abilities and the \emph{FIFA rank},
see Table~\ref{tab_rank}. 

%Table~\ref{tab_rank} compares the ranking of the 32 participating teams in the FIFA World Cup 2018 according to estimated abilities (left column), Elo rating (center column) and FIFA ranking (right column). The ranking according to the estimated abilities and the Elo ratings are very similar (Spearman correlation of $0.94$), while both have a smaller correlation with the FIFA ranking (Spearman correlation of $0.86$ and $0.90$, respectively).
%
%All three methods rank Germany and Brazil as the two top teams. Notable differences between the rankings can be seen, for example, for Spain and Belgium. Both the estimated abilities and the Elo rating rank Spain third while it is ranked ninth by FIFA. Belgium is ranked rather inhomogenously in positions 6, 8 and 3 by the different methods. More details on the comparison of estimated team abilities and the FIFA rank can be found in \citet{LeyWieEet2018}.

\begin{table}[!h]
\small
\caption{\label{tab_rank} Ranking of the participants of the UEFA EURO 2020 
according to estimated historic match abilities, bookmaker consensus abilities, 
average PM player rating and FIFA ranking.}\vspace{0.4cm}
\centering
\begin{tabular}{l|ll|ll|llll}
\multicolumn{1}{c}{}&\multicolumn{2}{c}{\textbf{Historic match}} &\multicolumn{2}{c}{\textbf{Bookmaker consensus}} &\multicolumn{2}{c}{\textbf{PM player}}  &\multicolumn{2}{c}{\textbf{FIFA}}\\
\multicolumn{1}{c}{}&\multicolumn{2}{c}{\textbf{abilities}} &\multicolumn{2}{c}{\textbf{abilities}} &\multicolumn{2}{c}{\textbf{rating}} &\multicolumn{2}{c}{\textbf{ranking}}\\
\toprule
% latex table generated in R 4.0.3 by xtable 1.8-4 package
% Mon Jun  7 17:12:26 2021
 1 & \includegraphics[width=0.4cm]{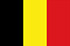} & Belgium & \includegraphics[width=0.4cm]{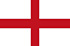} & England & \includegraphics[width=0.4cm]{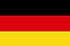} & Germany & \includegraphics[width=0.4cm]{BEL.png} & Belgium \\ 
  2 & \includegraphics[width=0.4cm]{ESP.png} & Spain & \includegraphics[width=0.4cm]{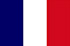} & France & \includegraphics[width=0.4cm]{ENG.png} & England & \includegraphics[width=0.4cm]{FRA.png} & France \\ 
  3 & \includegraphics[width=0.4cm]{ENG.png} & England & \includegraphics[width=0.4cm]{BEL.png} & Belgium & \includegraphics[width=0.4cm]{FRA.png} & France & \includegraphics[width=0.4cm]{ENG.png} & England \\ 
  4 & \includegraphics[width=0.4cm]{FRA.png} & France & \includegraphics[width=0.4cm]{GER.png} & Germany & \includegraphics[width=0.4cm]{ESP.png} & Spain & \includegraphics[width=0.4cm]{POR.png} & Portugal \\ 
  5 & \includegraphics[width=0.4cm]{POR.png} & Portugal & \includegraphics[width=0.4cm]{POR.png} & Portugal & \includegraphics[width=0.4cm]{BEL.png} & Belgium & \includegraphics[width=0.4cm]{ESP.png} & Spain \\ 
  6 & \includegraphics[width=0.4cm]{GER.png} & Germany & \includegraphics[width=0.4cm]{ESP.png} & Spain & \includegraphics[width=0.4cm]{POR.png} & Portugal & \includegraphics[width=0.4cm]{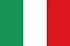} & Italy \\ 
  7 & \includegraphics[width=0.4cm]{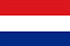} & Netherlands & \includegraphics[width=0.4cm]{ITA.png} & Italy & \includegraphics[width=0.4cm]{NED.png} & Netherlands & \includegraphics[width=0.4cm]{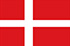} & Denmark \\ 
  8 & \includegraphics[width=0.4cm]{ITA.png} & Italy & \includegraphics[width=0.4cm]{NED.png} & Netherlands & \includegraphics[width=0.4cm]{ITA.png} & Italy & \includegraphics[width=0.4cm]{GER.png} & Germany \\ 
  9 & \includegraphics[width=0.4cm]{DEN.png} & Denmark & \includegraphics[width=0.4cm]{DEN.png} & Denmark & \includegraphics[width=0.4cm]{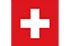} & Switzerland & \includegraphics[width=0.4cm]{SUI.png} & Switzerland \\ 
  10 & \includegraphics[width=0.4cm]{SUI.png} & Switzerland & \includegraphics[width=0.4cm]{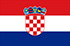} & Croatia & \includegraphics[width=0.4cm]{DEN.png} & Denmark & \includegraphics[width=0.4cm]{CRO.png} & Croatia \\ 
  11 & \includegraphics[width=0.4cm]{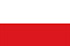} & Poland & \includegraphics[width=0.4cm]{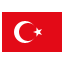} & Turkey & \includegraphics[width=0.4cm]{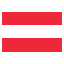} & Austria & \includegraphics[width=0.4cm]{NED.png} & Netherlands \\ 
  12 & \includegraphics[width=0.4cm]{CRO.png} & Croatia & \includegraphics[width=0.4cm]{SUI.png} & Switzerland & \includegraphics[width=0.4cm]{CRO.png} & Croatia & \includegraphics[width=0.4cm]{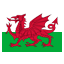} & Wales \\ 
  13 & \includegraphics[width=0.4cm]{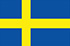} & Sweden & \includegraphics[width=0.4cm]{RUS.png} & Russia & \includegraphics[width=0.4cm]{TUR.png} & Turkey & \includegraphics[width=0.4cm]{SWE.png} & Sweden \\ 
  14 & \includegraphics[width=0.4cm]{RUS.png} & Russia & \includegraphics[width=0.4cm]{SWE.png} & Sweden & \includegraphics[width=0.4cm]{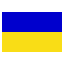} & Ukraine & \includegraphics[width=0.4cm]{POL.png} & Poland \\ 
  15 & \includegraphics[width=0.4cm]{WAL.png} & Wales & \includegraphics[width=0.4cm]{POL.png} & Poland & \includegraphics[width=0.4cm]{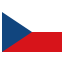} & Czech Republic & \includegraphics[width=0.4cm]{AUT.png} & Austria \\ 
  16 & \includegraphics[width=0.4cm]{AUT.png} & Austria & \includegraphics[width=0.4cm]{UKR.png} & Ukraine & \includegraphics[width=0.4cm]{POL.png} & Poland & \includegraphics[width=0.4cm]{UKR.png} & Ukraine \\ 
  17 & \includegraphics[width=0.4cm]{TUR.png} & Turkey & \includegraphics[width=0.4cm]{AUT.png} & Austria & \includegraphics[width=0.4cm]{SWE.png} & Sweden & \includegraphics[width=0.4cm]{TUR.png} & Turkey \\ 
  18 & \includegraphics[width=0.4cm]{UKR.png} & Ukraine & \includegraphics[width=0.4cm]{CZE.png} & Czech Republic & \includegraphics[width=0.4cm]{RUS.png} & Russia & \includegraphics[width=0.4cm]{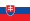} & Slovakia \\ 
  19 & \includegraphics[width=0.4cm]{CZE.png} & Czech Republic & \includegraphics[width=0.4cm]{WAL.png} & Wales & \includegraphics[width=0.4cm]{WAL.png} & Wales & \includegraphics[width=0.4cm]{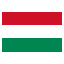} & Hungary \\ 
  20 & \includegraphics[width=0.4cm]{SVK.png} & Slovakia & \includegraphics[width=0.4cm]{HUN.png} & Hungary & \includegraphics[width=0.4cm]{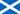} & Scotland & \includegraphics[width=0.4cm]{RUS.png} & Russia \\ 
  21 & \includegraphics[width=0.4cm]{SCO.png} & Scotland & \includegraphics[width=0.4cm]{SCO.png} & Scotland & \includegraphics[width=0.4cm]{SVK.png} & Slovakia & \includegraphics[width=0.4cm]{CZE.png} & Czech Republic \\ 
  22 & \includegraphics[width=0.4cm]{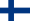} & Finland & \includegraphics[width=0.4cm]{FIN.png} & Finland & \includegraphics[width=0.4cm]{HUN.png} & Hungary & \includegraphics[width=0.4cm]{SCO.png} & Scotland \\ 
  23 & \includegraphics[width=0.4cm]{HUN.png} & Hungary & \includegraphics[width=0.4cm]{SVK.png} & Slovakia & \includegraphics[width=0.4cm]{FIN.png} & Finland & \includegraphics[width=0.4cm]{FIN.png} & Finland \\ 
  24 & \includegraphics[width=0.4cm]{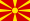} & North Macedonia & \includegraphics[width=0.4cm]{NMD.png} & North Macedonia & \includegraphics[width=0.4cm]{NMD.png} & North Macedonia & \includegraphics[width=0.4cm]{NMD.png} & North Macedonia \\ 
  
\end{tabular}
\end{table}

\subsection{Probabilities for UEFA EURO 2020 winner} \label{sec:simul}
In this section, the hybrid cforest model is applied to (new) data for the 
UEFA EURO 2020 (in advance of the tournament) to predict winning probabilities 
for all teams and to predict the tournament course.

The Poisson abilities were estimated by a bivariate Poisson 
model with a half period of  3 years (i.e.\ 1095 days). 
All matches of the 282 national teams played since 2003-05-27 up to 2021-05-26 are used for 
the estimation, what results in a total of 6953 matches.
All further predictor variables are taken as the latest values shortly before the UEFA EURO 
(and using the final squads of 26 players for all nations).
The bookmaker consensus abilities are based on the average odds of 19 bookmakers.

Note that due to the fact that the upcoming UEFA EURO tournament will not have a single hosting 
country, but the matches are distributed all over Europe, we believe that a potential home effect will
be much less pronounced compared to earlier tournaments, as also the average travel distances for
teams and fans should decrease. Moreover, due to the ongoing COVID-19 pandemic, the numbers 
of spectators allowed are substantially reduced compared to earlier tournaments. For these reasons,
we decided to set the {\it home} and {\it neighbor} dummy variables for all 24 teams equal to zero.

For each match in the UEFA EURO 2020, the hybrid cforest can 
be used to predict an expected number of goals for both teams. Given the expected 
number of goals, a real result is drawn by assuming two (conditionally) independent 
Poisson distributions for both scores. Based on these results, all 36 matches from the 
group stage can be simulated and final group standings can be calculated. Due to 
the fact that the full score-lines are simulated, we can precisely follow the official UEFA rules when
determining the final group standings (see again Footnote~\ref{fifa:rules}). %\footnote{The final group standings are determined by (1) the number of
%points, (2) the goal difference and (3) the number of scored goals.
%If several teams coincide with respect to all of these three criteria, a
%separate table is calculated based on the matches between the coinciding
%teams only. Here, again the final standing of the teams is
%determined following criteria (1)--(3). If still no distinct decision can
%be taken, the decision is induced by lot.\label{fifa:rules}}. 
This enables us to 
determine the matches in the round-of-sixteen and we can continue by 
simulating the knockout stage. In the case of draws in the knockout stage, 
we simulate extra-time by a second simulated result. However, here we multiply 
the expected number of goals by the factor $1/3$ to account for the shorter time 
to score (30 min instead of 90 min). In  the case of a further draw in extra-time 
we simulate the penalty shootout by a (virtual) coin flip.

Following this strategy, a whole tournament run can be simulated, which we repeat 100,000
times. Based on these simulations, for each of the 24 participating
teams probabilities to reach the single knockout stages and,
finally, to win the tournament are obtained. These are summarized
in Table~\ref{winner_probs} together with the (average) winning probabilities
based on 19 different bookmakers for comparison.

\begin{table}[!h]
\small
\caption{\label{winner_probs}Estimated probabilities (in \%) for reaching the 
different stages in the UEFA EURO 2020 for all 24 teams based on 100,000 
simulation runs of the UEFA EURO 2020 together with (average) winning probabilities 
based on the odds of 19 bookmakers.}\vspace{0.4cm}

\centering
% latex table generated in R 4.0.3 by xtable 1.8-4 package
% Sun Jun  6 21:10:28 2021
\begin{tabular}{lllrrrrrr}
   \toprule
 &  &  & Round & Quarter & Semi & Final & European & Bookmakers \\ 
   &  &  & of 16 & finals & finals &  & Champion & consensus \\ 
   \midrule
1. & \includegraphics[width=0.4cm]{FRA.png} & FRA & 89.7 & 59.5 & 39.7 & 25.0 & 14.8 & 15.0 \\ 
  2. & \includegraphics[width=0.4cm]{ENG.png} & ENG & 94.6 & 56.3 & 35.3 & 22.9 & 13.5 & 14.8 \\ 
  3. & \includegraphics[width=0.4cm]{ESP.png} & ESP & 94.0 & 66.8 & 35.4 & 21.9 & 12.3 & 9.9 \\ 
  4. & \includegraphics[width=0.4cm]{POR.png} & POR & 85.3 & 52.3 & 31.7 & 18.6 & 10.1 & 9.0 \\ 
  5. & \includegraphics[width=0.4cm]{GER.png} & GER & 85.3 & 52.3 & 32.5 & 18.8 & 10.1 & 9.6 \\ 
  6. & \includegraphics[width=0.4cm]{BEL.png} & BEL & 91.5 & 54.6 & 32.2 & 16.2 & 8.3 & 12.1 \\ 
  7. & \includegraphics[width=0.4cm]{ITA.png} & ITA & 88.8 & 56.6 & 32.2 & 15.9 & 7.9 & 7.5 \\ 
  8. & \includegraphics[width=0.4cm]{NED.png} & NED & 93.4 & 52.1 & 28.3 & 13.0 & 6.1 & 6.5 \\ 
  9. & \includegraphics[width=0.4cm]{DEN.png} & DEN & 84.5 & 44.4 & 23.2 & 10.2 & 4.6 & 2.9 \\ 
  10. & \includegraphics[width=0.4cm]{CRO.png} & CRO & 78.0 & 36.8 & 16.3 & 7.4 & 3.1 & 2.4 \\ 
  11. & \includegraphics[width=0.4cm]{SUI.png} & SUI & 72.3 & 34.7 & 15.3 & 5.7 & 2.2 & 1.1 \\ 
  12. & \includegraphics[width=0.4cm]{AUT.png} & AUT & 80.9 & 33.2 & 13.5 & 4.6 & 1.5 & 0.8 \\ 
  13. & \includegraphics[width=0.4cm]{POL.png} & POL & 66.2 & 29.8 & 10.2 & 3.9 & 1.2 & 1.1 \\ 
  14. & \includegraphics[width=0.4cm]{SWE.png} & SWE & 59.8 & 25.6 & 8.7 & 3.2 & 1.0 & 0.9 \\ 
  15. & \includegraphics[width=0.4cm]{TUR.png} & TUR & 53.3 & 20.8 & 7.8 & 2.4 & 0.7 & 1.6 \\ 
  16. & \includegraphics[width=0.4cm]{WAL.png} & WAL & 53.7 & 20.6 & 7.4 & 2.1 & 0.6 & 0.6 \\ 
  17. & \includegraphics[width=0.4cm]{SCO.png} & SCO & 49.8 & 19.3 & 6.3 & 2.0 & 0.6 & 0.4 \\ 
  18. & \includegraphics[width=0.4cm]{RUS.png} & RUS & 52.0 & 16.8 & 5.3 & 1.4 & 0.4 & 0.9 \\ 
  19. & \includegraphics[width=0.4cm]{CZE.png} & CZE & 40.8 & 14.4 & 4.4 & 1.3 & 0.3 & 0.6 \\ 
  20. & \includegraphics[width=0.4cm]{UKR.png} & UKR & 57.4 & 16.9 & 5.1 & 1.3 & 0.3 & 1.0 \\ 
  21. & \includegraphics[width=0.4cm]{SVK.png} & SVK & 44.9 & 16.5 & 4.6 & 1.3 & 0.3 & 0.3 \\ 
  22. & \includegraphics[width=0.4cm]{FIN.png} & FIN & 37.1 & 9.3 & 2.3 & 0.5 & 0.1 & 0.2 \\ 
  23. & \includegraphics[width=0.4cm]{NMD.png} & NMD & 32.9 & 7.1 & 1.6 & 0.3 & 0.1 & 0.2 \\ 
  24. & \includegraphics[width=0.4cm]{HUN.png} & HUN & 13.9 & 3.4 & 1.0 & 0.2 & 0.0 & 0.2 \\ 
   \bottomrule
\end{tabular}

\end{table}

We can see that, according to our hybrid cforest model, 
the current FIFA World champion France is the favored team with a predicted winning probability of $14.8\%$ 
followed by England, Spain, Portugal and Germany. Overall, this result seems mostly in line with the probabilities 
from the bookmakers, as we can see in the last column. However, e.g.\ for Belgium, which the bookmakers rate on place three with
a winning chance of 12.1\%, the cforest model calculates substantially lower chances (8.3\%). 
%However, while the bookmakers only slightly favor England, 
%the xgboost model predicts a more clear advantage. 
Beside the probabilities of becoming European champion, Table~\ref{winner_probs} provides some 
further interesting insights also for the single stages within the tournament. 
For example, it is interesting to see that while Belgium has a much higher probability to reach the 
round of 16 than Germany ($91.5\%$ vs.\ $85.3\%$), Germany has higher 
chances to reach the final ($16.2\%$ vs.\ $18.8\%$) and become European champion ($8.3\%$ vs.\ $10.1\%$). 
This is probably related to the fact
that German is assigned to a very tough group with France and Portugal.
%the two favored teams USA and France have similar chances to at least reach the 
%round-of-sixteen ($98.4\%$ and $95.9\%$, respectively), while the probabilities to at least reach 
%the quarter finals differ significantly. 
%While USA has a probability of $75.5\%$ to reach at least the quarter finals 
%and of $53.4$ to reach at least the semi finals, France only achieves respective probabilities of 
%$66.8\%$ and $40.7$. Obviously, in contrast to USA, France has 
%a rather high chance to meet a strong opponent in the round-of-sixteen and the quarter finals. 
It is also interesting that some teams have a fair chance to reach the round of 16, but then their chances decrease drastically to 
reach the next round. Ukraine, for example, has a moderate chance of $57.4\%$ to reach the knockout stage, but only 
a rather low chance of $16.9\%$ the reach the quarter finals.

The lowest chances are given to Hungary, though they do not have the weakest squad 
(see Table~\ref{tab_rank}). The reason is their very low chance to attain the Round of 16 because of 
their bad luck with the draw: they are in the strongest group with France, Germany and Portugal

\section{Concluding remarks}\label{sec:conclusion}

In this work, we proposed hybrid modeling approaches for the scores of
international football matches which combine random forests, an extreme gradient boosting approach 
and lasso-penalized Poisson regression with several different ranking methods, namely
a current ability ranking based on historic matches, abilities based on bookmakers' odds and plus-minus player rankings.
While the machine learning models are principally based on the competing teams' covariate information,
the latter  components provide ability parameters, which serve as adequate 
estimates of the current team strengths a well as of the information contained in the bookmakers' odds. 
In order to combine the methods, the current ability and PM player ranking methods need to be repeatedly applied 
to historical match data preceding each UEFA EURO from the training data. This way, for each 
UEFA EURO in the training data and each participating team ability estimates are obtained. 
Similarly, the bookmaker consensus abilities are obtained by inverse tournament simulation
based on the aggregated winning odds from several online bookmakers.
These rankings and ability estimates can be added as additional covariates to the set of covariates used 
in the machine learning procedures. We compared the predictive performances of
the approaches in a leave-one-tournament-out competition on the data of the four 
preceding UEFA EUROs 2004-2016 and found the highest potential in the
cforest model, closely followed by the xgboost model.

Additionally, based on the estimates of the hybrid xgboost model 
on the training data, we repeatedly simulated the upcoming UEFA EURO 2020 100,000 times. 
According to the simulations, the current World champion France with a winning probability of 14.8\% is the top 
favorite for winning the title, followed by England (13.5\%) and Spain (12.3\%).
Furthermore, survival probabilities for all teams and at all tournament stages are
provided. 

Even though in our prediction competition in Section~\ref{sec:combine} the {\it extreme gradient boosting} 
technique was not yet able to substantially outperform the other approaches, though being well-known 
in the machine learning community for its high predictive power, we believe that this is partly due to the 
small sample size of our training data of the four UEFA EUROs 2004-2016 (and in the competition the training 
data size even decreased due to the leave-one-tournament-out strategy). 
Compared to the other methods, xgboost involves a more sophisticated tuning process with several tuning 
parameters and, hence, to fully exploit its high potential the more training data are available the better.
On our small training data sets, however, we experienced a rather high sensitiveness and instability during the tuning process.
Nevertheless, we think that our results already indicate that the method is very promising
for modeling and predicting football matches. We are looking forward with excitement to comparing
the cforest and xgboost approach again in an extensive ex-post analysis, when the UEFA EURO 2020 is finished.

\subsection*{Acknowledgment}
We thank Jonas Heiner for his tremendous effort in helping us to collect
the covariate data.

\pagebreak

%%%%%%%%%%%%
%%%%%%%%%%%%
\appendix
\section*{Appendix}

\section{Some notations and definitions}\label{sec:general}

Kronecker's delta,  which is used in  Section~\ref{sec:combine} in the formula of the multinomial likelihood and the RPS, is defined as follows:

$$
\delta_{ij} = \begin{cases}1,\quad \text{if}\,\, i=j\,,\\
0,\quad \text{otherwise}\,.
\end{cases}
$$
\medskip

\noindent The Skellam distribution, which is also used in Section~\ref{sec:combine}, is the discrete probability distribution of the integer random variable that is defined as the difference $K:=Y_1-Y_2$ of two independent Poisson distributed random variables $Y_1, Y_2$ with respective event rates $\lambda_1,\lambda_2$. The corresponding probability mass function is given by
$$
P(K=k)=e^{-(\lambda_1+\lambda_2)}\left(\frac{\lambda_1}{\lambda_2}\right)^{k/2}I_k(2\sqrt{\lambda_1\lambda_2}),\quad k \in \mathbb{Z},
$$
where $I_k(\cdot)$ is the modified Bessel function of the first kind (for more details, see \citealp{Ske:46}). Now let $Y_1$ and $Y_2$  denote the (conditionally independent) Poisson-distributed numbers of goals of two soccer teams competing in a match. Then, the three probabilities $P(Y_1>Y_2), P(Y_1=Y_2)$ and $P(Y_1<Y_2)$ can be easily obtained by computing
$P(K>0), P(K=0)$ and $P(K<0)$ via the Skellam distribution.

\section{Additional data material}\label{sec:appendix:ata}

\begin{table}[h!]
\footnotesize
\centering
\begin{tabular}{rrrrrrrrrrrrr}
\hline
 & FRA & ITA & NED & POR & ESP & GER & ENG & CZE \\ \hline
Oddset & 3.25 & 5 & 5.5 & 6 & 6.5 & 7 & 7 & 7 \\
\hline
 & SWE & DEN & RUS & GRE & CRO & BUL & SUI & LVA \\ \hline
Oddset & 20 & 20 & 40 & 45 & 45 & 60 & 60 & 100 \\\hline
\end{tabular}
\caption{\label{tab:odds2004} Quoted odds from ODDSET for the 16~teams in the EURO~2004.}
\end{table}

\pagebreak

\begin{table}[h!]
%\footnotesize
\scriptsize
\centering
\begin{tabular}{rrrrrrrrrrrrr}
\hline
 & FRA & ENG & BEL & ESP & GER & POR & ITA & NED \\ \hline
bwin & 5.5 & 6.0 & 7.50 &  9.0 &  8.0 &  9.00 & 11.00 & 13.0 \\
bet365 & 5.5 & 6.0 & 7.00 &  8.5 &  8.0 &  9.00 & 12.00 & 13.0 \\
Sky Bet & 6.0 & 6.0 & 7.00 &  9.0 &  9.0 &  9.00 & 12.00 & 12.0 \\
Paddy Power & 6.0 & 5.0 & 7.50 &  8.0 &  9.0 & 11.00 &  9.00 & 13.0 \\
William Hill & 5.5 & 6.0 & 7.00 &  8.5 &  9.0 &  9.00 & 12.00 & 13.0 \\
Betfair Sportsbook & 6.0 & 5.0 & 7.50 &  8.0 &  9.0 & 11.00 &  9.00 & 13.0 \\
Bet Victor & 5.5 & 5.5 & 7.00 &  9.0 & 10.0 &  9.00 & 11.00 & 13.0 \\
Unibet & 6.0 & 6.5 & 7.00 & 10.0 & 10.0 &  9.00 & 13.00 & 15.0 \\
Mansion Bet & 6.0 & 6.0 & 7.00 &  8.5 &  8.0 & 10.00 & 12.00 & 12.0 \\
Smarkets Sportsbook & 6.0 & 6.0 & 7.60 &  9.8 & 10.0 & 10.00 & 12.50 & 15.5 \\
Betway & 6.0 & 6.0 & 7.00 &  8.5 &  9.0 & 10.00 & 12.00 & 13.0 \\
Boylesports & 5.5 & 6.0 & 7.00 &  8.5 &  8.5 &  9.00 & 12.00 & 13.0 \\
10Bet & 6.0 & 6.0 & 7.00 &  8.5 &  8.5 &  9.00 & 12.00 & 12.0 \\
Sport Nation & 5.5 & 5.5 & 7.00 &  8.5 & 10.0 &  8.00 & 11.00 & 12.0 \\
Vbet & 5.5 & 5.9 & 6.80 &  8.5 &  8.0 &  8.80 & 12.00 & 13.0 \\
Sporting Index & 6.0 & 6.0 & 6.50 &  8.0 &  8.5 & 10.00 & 12.00 & 13.0 \\
RedZone & 6.0 & 6.0 & 7.75 &  8.5 &  8.5 &  8.75 & 10.75 & 12.0 \\
Spreadex & 5.5 & 5.5 & 6.50 &  8.5 &  9.0 & 10.00 & 10.00 & 13.0 \\
Smarkets & 5.8 & 6.0 & 7.80 &  9.6 &  9.8 & 10.80 & 12.20 & 15.2 \\
\hline
 & DEN & CRO & TUR & SUI & POL & UKR & SWE & RUS \\ \hline
bwin & 29 & 34 & 41 & 81 &  81 &  81 & 101 &  81 \\
bet365 & 29 & 34 & 51 & 67 &  81 &  51 & 101 &  67 \\
Sky Bet & 34 & 29 & 51 & 67 &  81 &  67 &  81 &  51 \\
Paddy Power & 26 & 31 & 67 & 91 &  67 &  91 &  91 &  76 \\
William Hill & 29 & 34 & 51 & 81 &  81 & 101 &  81 & 101 \\
Betfair Sportsbook & 26 & 31 & 67 & 91 &  67 &  91 &  91 &  76 \\
Bet Victor & 26 & 41 & 51 & 81 &  81 & 126 &  81 & 126 \\
Unibet & 34 & 41 & 61 & 71 &  81 & 101 & 101 & 101 \\
Mansion Bet & 31 & 31 & 51 & 67 &  67 &  81 &  67 & 101 \\
Smarkets Sportsbook & 34 & 44 & 65 & 90 & 100 & 100 & 120 & 240 \\
Betway & 29 & 34 & 51 & 81 &  81 &  67 & 101 &  80 \\
Boylesports & 29 & 34 & 51 & 67 &  81 &  81 &  81 &  81 \\
10Bet & 31 & 34 & 51 & 67 &  81 &  81 &  81 &  81 \\
Sport Nation & 29 & 34 & 51 & 67 &  81 &  81 &  81 &  81 \\
Vbet & 29 & 34 & 51 & 67 &  81 &  51 & 101 &  67 \\
Sporting Index & 29 & 34 & 61 & 67 &  67 &  81 & 101 & 101 \\
RedZone & 29 & 34 & 34 & 67 &  81 &  81 &  81 &  81 \\
Spreadex & 29 & 29 & 51 & 51 &  81 &  81 &  81 & 126 \\
Smarkets & 33 & 43 & 64 & 88 &  98 &  98 & 118 & 235 \\
\hline
 & AUT & CZE & WAL & SCO & SVK & FIN & HUN & MKD \\ \hline
bwin & 101 & 151 & 101 & 151 & 201 & 201 & 201 & 201 \\
bet365 &  81 & 151 & 201 & 251 & 251 & 501 & 401 & 501 \\
Sky Bet & 101 & 151 & 101 & 151 & 151 & 151 & 251 & 501 \\
Paddy Power &  91 & 126 & 126 & 251 & 326 & 326 & 326 & 501 \\
William Hill & 126 & 126 & 126 & 251 & 251 & 251 & 401 & 501 \\
Betfair Sportsbook &  91 & 126 & 126 & 251 & 326 & 326 & 326 & 501 \\
Bet Victor & 126 & 126 & 126 & 251 & 501 & 501 & 501 & 501 \\
Unibet & 201 & 151 & 101 & 301 & 301 & 501 & 801 & 501 \\
Mansion Bet & 101 & 101 & 126 & 201 & 301 & 401 & 401 & 501 \\
Smarkets Sportsbook & 200 & 190 & 260 & 300 & 300 & 500 & 500 & 500 \\
Betway &  80 & 101 & 151 & 251 & 501 & 501 & 501 & 501 \\
Boylesports & 101 & 101 & 101 & 201 & 251 & 251 & 401 & 501 \\
10Bet &  81 & 126 & 151 & 201 & 201 & 401 & 401 & 501 \\
Sport Nation &  81 & 126 & 151 & 201 & 201 & 401 & 401 & 501 \\
Vbet &  81 & 151 & 201 & 251 & 251 & 501 & 401 & 501 \\
Sporting Index & 101 & 151 & 151 & 251 & 251 & 301 & 301 & 501 \\
RedZone &  81 & 126 & 151 & 201 & 201 & 401 & 401 & 501 \\
Spreadex & 151 &  81 & 151 & 251 & 501 & 401 & 501 & 401 \\
Smarkets & 196 & 186 & 255 & 294 & 294 & 490 & 490 & 490 \\\hline
\end{tabular}
\caption{\label{tab:odds2020} Quoted odds from 19~online bookmakers
for the 24~teams in the EURO~2020 obtained on 2021-05-31 from
\url{https://www.oddschecker.com/} and \url{https://www.bwin.com/}, respectively.}
\end{table}

\bibliographystyle{apalike}
\bibliography{literatur}

\end{document}